\documentclass[runningheads]{llncs}

\usepackage{eccv}

\usepackage{eccvabbrv}

\usepackage{graphicx}
\usepackage{booktabs}
\usepackage{pifont}
\usepackage[ruled, linesnumbered, noline]{algorithm2e}
\usepackage{float} 
\usepackage{nicematrix}
\usepackage{multirow}

\usepackage[accsupp]{axessibility}  

\usepackage{hyperref}

\usepackage{orcidlink}

\begin{document}

\title{TEASR: Training-Efficient Any-Step Diffusion Transformer for Real-World Image Super-Resolution} 

\titlerunning{TEASR}

\author{Xiang Gao\orcidlink{0009-0003-6291-9522}, 
Chenxin Zhu\orcidlink{0009-0002-0679-6022}, 
Yushun Fang\orcidlink{0009-0008-8105-1083},\\ 
Qiang Hu\thanks{Corresponding author.}\orcidlink{0000-0003-4645-9776}, 
Xiaoyun Zhang\orcidlink{0000-0001-7680-4062} \\[3pt] 
}

\authorrunning{X.~Gao et al.}

\institute{Shanghai Jiao Tong University, China \\
\email{\{frxg0918, Zcx-sjtu, ethanfang, qiang.hu, xiaoyun.zhang\}@sjtu.edu.cn}}

\maketitle

\begin{figure}[!h]
    \centering
    \includegraphics[width=1.0\linewidth]{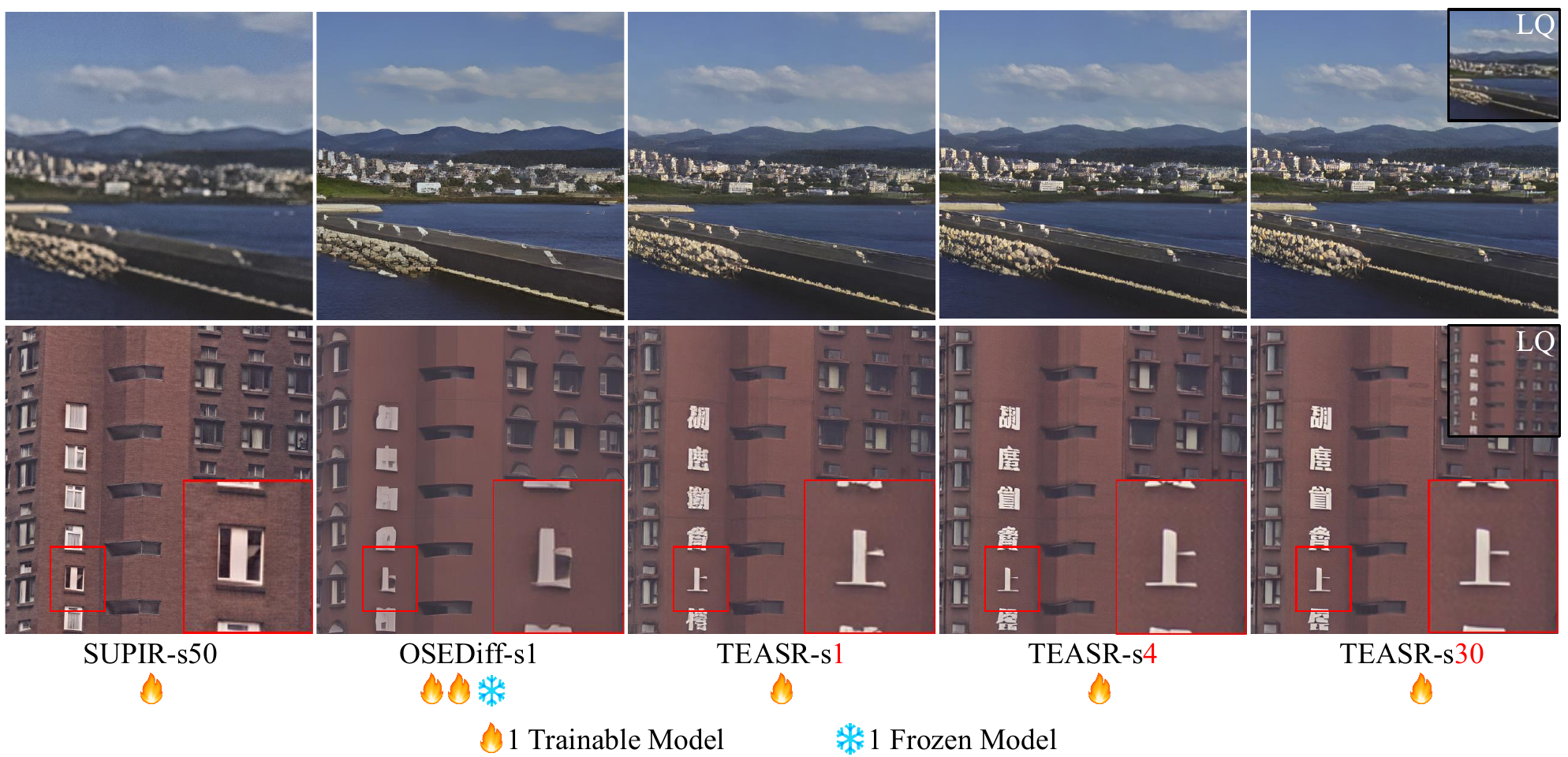}
    \vspace{-10pt}
    \caption{Qualitative comparison of our any-step TEASR against purely multi-step and one-step method. 's' denotes the sampling steps. \textcolor{red}{Red} represents any-step choice. While multi-step methods require iterative refinement and single-step approaches suffer from heavy training overhead, our approach achieves satisfying quality at one-step and delivers even more superior performance as inference steps increasing.}
    \vspace{-25pt}
    \label{fig:teaser}
\end{figure}
\begin{abstract}
  Diffusion models excel in Real-World Image Super-Resolution (Real-ISR) due to their powerful generative priors but suffer from slow iterative sampling. Although existing one-step distillation methods accelerate inference, they typically require auxiliary teacher models that inflate training memory and restrict scalability to large-scale architectures. Furthermore, these fixed-step models lack the flexibility to trade off speed for quality. In this paper, we propose TEASR, a training-efficient any-step diffusion framework for Real-ISR that enables both one-step and multi-step restoration within a unified model. Our key idea is to perform self-adversarial distillation within a single diffusion model, eliminating the need for auxiliary teachers or discriminators. Specifically, we propose a timestep-aware rectification strategy that stabilizes one-step generation across noise levels. These two designs further enables the distillation of 20B-parameter diffusion models on a single GPU, significantly improving training efficiency. Moreover, we introduce a dual-branch diffusion transformer with decoupled timestep condition to separate the current noise state and the denoising target to enhance sampling quality. Extensive experiments demonstrate that TEASR supports seamless any-step sampling and consistently outperforms state-of-the-art methods across multiple datasets. Code: {\url{https://github.com/MediaX-SJTU/TEASR}}.
\end{abstract}

\section{Introduction}
\label{sec:intro}

Recently diffusion models~\cite{ho2020denoising, rombach2022high, peebles2023scalable, labs2025flux, wu2025qwenimagetechnicalreport, shen2026agentic, wu2025megafusion} have achieved remarkable success in image generation and various downstream vision tasks~\cite{zhang2025td, zhang2025lato, fang2025robust, fang2025one, liu2025moa}. In particular, diffusion-based approaches have established new state-of-the-art performance in Real-ISR by generating highly realistic details. However, their iterative sampling process~\cite{wang2024exploiting, lin2024diffbir, yu2024scaling, duan2025dit4sr} significantly hinders real-time deployment.

To accelerate diffusion models, recent researches~\cite{salimans2022progressive, song2023consistency, frans2025one} have focused on advanced samplers and few-step distillation. Despite their success, these methods still face a steep quality drop-off in extreme one-step sampling scenarios. Emerging one-step distillation techniques, such as Distribution Matching Distillation(DMD)~\cite{yin2024one} and adversarial distillation training~\cite{sauer2024adversarial, lin2025diffusion}, have achieved satisfying one-step results. Recent diffusion-based SR methods have adopt some distillation methods to achieve one-step sampling. As shown in \cref{fig:teaser}, multi-step super-resolution models(e.g., SUPIR~\cite{yu2024scaling}) exhibit training efficiency and stability, as they require only one single trainable model, but they cannot sample with few-step even one-step. One-step super-resolution models such as OSEDiff~\cite{wu2024one} can infer quickly, but they usually maintain multiple models during training, raising memory consumption. Besides one-step methods cannot adjust the number of inference steps, abandoning the ability to increase sampling steps for more natural results. But our TEASR can perform any-step sampling, ranging from one-step to typically 30-step without extra training. As the inference steps increase, our method generates more natural restoration results.

To conclude, current one-step SR methods introduce two critical challenges in both training and application. 
Firstly, these methods suffer from prohibitive training overhead because they necessitate maintaining multiple extra teacher models or discriminators to provide distillation signals, which significantly limits training efficiency and restricts their application to large-scale architectures. Secondly, these one-step models are typically confined to a fixed, preset sampling paradigm, thereby locking the restoration quality to a single forward pass and eliminating the ability to adjust inference costs according to practical needs.


%
To address these challenges, we propose \textbf{TEASR} (\textbf{T}raining-\textbf{E}fficient \textbf{A}ny-\textbf{S}tep Diffusion Transformer for Real-World Image \textbf{S}uper-\textbf{R}esolution), a minimalist framework that enables training-efficient and {any-step} Real-ISR under a unified conditional flow-matching formulation. TEASR performs {Self-Adversarial Distillation (SAD)} entirely within a single model, eliminating auxiliary teacher models or discriminators and substantially reducing training overhead, which makes it feasible to fine-tune large-scale backbones (e.g., the 20B-parameter Qwen-Image~\cite{wu2025qwenimagetechnicalreport}) on a single NVIDIA A100 GPU. Within SAD, we propose a simple but effective Time-Aware Rectification(TAR) mechanism which adjusts the rectification strength according to the noise level, thus providing more precise signals to better align the restored results with real data distribution.


Our framework introduces a dual-branch design for DiT architecture with Decoupled Timestep Condition (DTC) that explicitly separates the generative prior from conditional guidance. In traditional self-consistent distillation~\cite{song2023consistency, frans2025one}, a second timestep is introduced to represent the flow destination. By making the noise branch conditioned on the current timestep $t_{\text{cur}}$ and the condition branch conditioned on the target timestep $t_{\text{tar}}$, TEASR preserves fine-grained noise-level information in self-consistent objective, enabling the model to precisely reason about the instantaneous noise state and the intended destination jointly without information loss. This design eliminates information entanglement, thereby ensuring superior restoration fidelity in any-step sampling, providing a practical adjustment to trade inference efficiency for restoration quality.

To summarize, our contributions are as follows:
\begin{itemize}
    \item We propose TEASR, a training-efficient any-step SR method that utilizes Self-Adversarial Distillation(SAD) to eliminate the need for auxiliary teachers or discriminators during training. As a core refinement within this framework, we introduce Time-Aware Rectification(TAR) to adaptively weight the distillation signals, ensuring the restoration results are more precisely aligned with the real data distribution. 
    \item We introduce a dual-branch DiT architecture with Decoupled Timestep Condition(DTC). This design decouples the two timesteps from the self-consistent objective, assigning them separately to the noise and condition branches to ensure that both timestep information is fully preserved within the attention blocks to achieve more faithful reconstruction in any-step sampling.
    \item Extensive experiments demonstrate that TEASR supports seamless any-step sampling and consistently outperforms state-of-the-art methods across multiple datasets. Our any-step sampling have good performance scalability, enabling a boost in restoration quality as the number of inference steps increases.
\end{itemize}

\section{Related Work} 


\subsection{Real-World Image Super-Resolution}

Real-world image super-resolution(Real-ISR) aims to recover high-quality(HQ) images from low-quality(LQ) inputs suffering from unknown and complex degradations. Early approaches predominantly relied on {pixel-wise regression} objectives (e.g., PSNR-oriented methods). While effective for full-reference pixel-wise metric scores, these methods often produce over-smoothed results lacking perceptual realism due to the ill-posed nature of the problem. To address this, {Generative Adversarial Networks(GANs)}~\cite{GAN, wang2021real} were introduced to model the data distribution, yielding visually pleasing textures. GAN-based method such as Real-ESRGAN~\cite{wang2021real} has made significant progress. However, GANs frequently suffer from training instability, mode collapse, and artifacts, limiting their reliability and robustness in practical applications.

Recently, {diffusion models}~\cite{ho2020denoising, rombach2022high, peebles2023scalable, labs2025flux, wu2025qwenimagetechnicalreport} have emerged as the new state-of-the-art in Real-ISR, surpassing GANs in both training stability and generative modeling ability, achieving visually more realistic performance. Methods like~\cite{wang2024exploiting, lin2024diffbir, yu2024scaling, duan2025dit4sr} leverage powerful pretrained diffusion priors to handle real-world degradations effectively. More recently, diffusion transformers(DiTs)~\cite{peebles2023scalable} have shown great potential in scaling capabilities of diffusion models. DiT4SR~\cite{duan2025dit4sr} adopts pretrained large-scale DiT models and finetune them to Real-ISR task, achieving superior perceptual realism in restoration results. 

\subsection{Diffusion Distillation}

The iterative sampling process of diffusion models poses a significant bottleneck for real-time applications. Consequently, extensive research has focused on {diffusion distillation} to reduce sampling steps. 
Diffusion distillation aims to learn a direct mapping from noise to data (or between arbitrary time steps). Progressive distillation~\cite{salimans2022progressive} adopts a pre-trained teacher model to progressively distill a student model to fewer steps. {Consistency Models(CMs)}~\cite{song2023consistency, frans2025one} are optimized by enforcing self-consistency along the probability flow trajectory. While these methods successfully reduce sampling to few steps (e.g., 4--10 steps), they often struggle to maintain high quality when distilled to {one-step}.

To achieve true {one-step generation}, recent works have explored {distribution matching distillation(DMD)}-like and {adversarial distillation}. Methods such as DMD~\cite{yin2024one} and Adversarial Diffusion Distillation(ADD)~\cite{sauer2024adversarial} introduce auxiliary discriminators or teacher models to align the student's one-step output distribution with the real data distribution. Similarly, in the SR domain, methods like OSEDiff~\cite{wu2024one} and HYPIR~\cite{lin2025harnessing} have adopted multi-model distillation strategies to achieve one-step super-resolution. 
{However, two critical limitations persist:} these state-of-the-art one-step frameworks typically require maintaining {multiple co-resident models} (e.g., teacher, student, and discriminator) during training. This significantly increases GPU memory consumption (often requiring about 2--3$\times$ the memory of a single model), making it prohibitively expensive to distill large-scale foundation models like Qwen-Image or Flux~\cite{labs2025flux}. Furthermore, most existing one-step SR methods~\cite{wu2024one, dong2025tsd, fang2025one} are fixed to a specific one-step sampling paradigm, losing the ability to leverage additional computation for higher quality when latency permits.

While recent works~\cite{cheng2025twinflow, peng2025facm} have demonstrated the potential of self-adversarial distillation within a unified model for text-to-image generation, directly adapting this paradigm to the Real-ISR task is highly non-trivial. Unlike text-to-image generation, Real-ISR strictly demands both reconstruction fidelity and perceptual realism. Furthermore, precise perception of noise levels is also essential for Real-ISR to avoid over-smoothing while effectively removing degradations~\cite{fang2025one}. To handle this domain shift and address the open challenge of enabling {any-step} controllability, TEASR significantly advances the unified distillation framework for Real-ISR.

\section{Method}
\textbf{Consistency Models(CMs)}~\cite{song2023consistency} aim to enforce self-consistency across the Probability Flow ODE (PF-ODE) trajectory. Advanced frameworks~\cite{frans2025one} generalize this property by introducing a target timestep $t_{\text{tar}}$(here we use $s$ for simplicity), which acts as an explicit destination for each sampling step. This formulation defines a velocity-based consistency function $v_\theta(\mathbf{z}_t, t, s)$ that predicts the transition from current state ${z}_t$ at time $t$ to a targeted state at time $s$. The core objective is to  ensure that a direct step from $t$ to $s$ remains consistent with an iterative path passing through an intermediate state. Formally, let $t, s$ be the current and target timesteps respectively, and $m = \frac{t+s}{2}$ be the midpoint. The generalized self-consistent objective is defined as:
\begin{gather}
\mathbf{v}_1 = v_\theta({z}_t, t, s) \\
{z}_m = {z}_t - \frac{t-s}{2} v_\theta({z}_t, t, m) \\
\mathbf{v}_2 = \frac{1}{2} v_\theta({z}_t, t, m) + \frac{1}{2} v_\theta({z}_m, m, s) \\
{L}_{\text{cons}} = || \mathbf{v}_1 - \mathbf{v}_2 ||_2
\end{gather}
where $\mathbf{v}_1$ represents the single-step shortcut, and $\mathbf{v}_2$ denotes the integrated velocity of the two-step composition. In this paradigm, $s$ is no longer a passive parameter but a crucial guidance signal that anchors the denoising direction. Without the explicit conditioning on $s$, the model would suffer from temporal ambiguity, failing to distinguish between trajectories with different refinement destinations. This dual-timestep dependency $(t, s)$ is essential for enabling any-step inference but introduces the challenge of interference or entanglement between two temporal signals, which we will address in \cref{sec:DTC}.

\begin{figure}[tbp]
    \centering
    \includegraphics[width=1.0\linewidth]{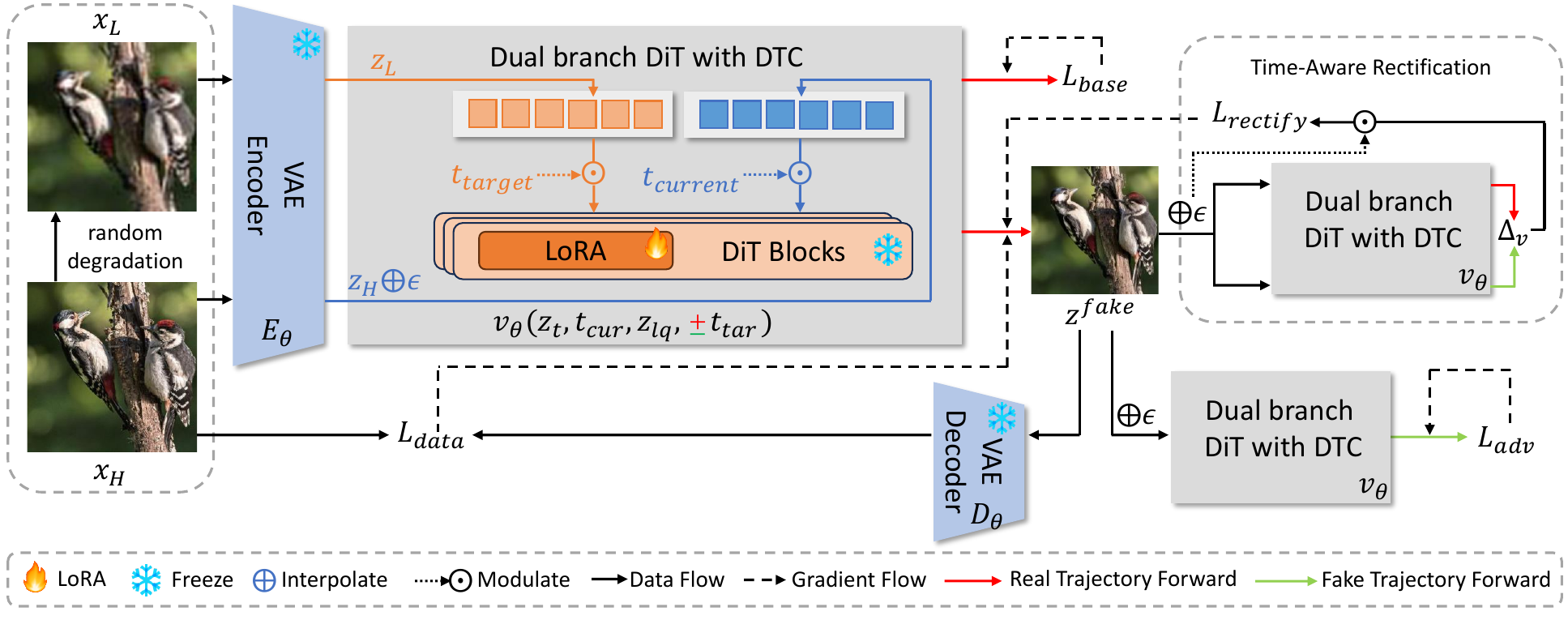}
    \caption{\textbf{Overall SAD Framework.} We propose a dual-branch DiT with DTC to introduce LQ condition and decouple two timesteps. For training, we first optimize the \textcolor{red}{real trajectory} with flow matching loss and self-consistent loss(\textcolor{black}{$L_{\text{base}}$}). We then estimate a fake image within one-step and decode it back to pixel domain to calculate its reconstruction loss(\textcolor{black}{$L_{\text{data}}$}). Then the \textcolor{green}{fake trajectory} learns the one-step generated image data distribution(\textcolor{black}{$L_{\text{adv}}$}). Finally we perform TAR with both \textcolor{red}{real} and \textcolor{green}{fake} trajectory to optimize the generation process of the one-step estimated image(\textcolor{black}{$L_{\text{rectify}}$}).}
    \label{fig:framework}
\end{figure}

\textbf{Framework overview.} \cref{fig:framework} illustrates the overall Self-Adversarial Distillation architecture of TEASR. Our framework operates within a unified conditional flow matching paradigm and introduces a Self-Adversarial Distillation scheme that eliminates the need for external teacher or discriminator models. The core design features a dual-branch DiT backbone with Decoupled Timestep Condition, where the noise branch is conditioned on the current timestep($t_{\text{cur}}$) and the condition branch is conditioned on the target timestep($t_{\text{tar}}$). Besides we propose a Time-Aware Rectification to precisely provide distillation signals to better straighten the generation path.



\subsection{Self-Adversarial Distillation for SR}
\textbf{Twin trajectories for Self-Adversarial Distillation.} Inspired by recent works~\cite{peng2025facm, cheng2025twinflow}, we introduce twin trajectories in one model to create a self-adversarial signal, thus eliminating the need for extra models.

Firstly, we train the model to serve as both an instantaneous velocity predictor and a self-consistent path-integral predictor by minimizing a standard flow-matching objective together with a self-consistent regularizer:
\begin{gather}
    L_{\text{base}} = L_{\text{flow}} + \lambda_{\text{cons}}\, L_{\text{cons}}
    \label{eq:loss_base}
\end{gather}
where $L_{\text{flow}}$ enforces accurate instantaneous velocity estimation, and $L_{\text{cons}}$ encourages consistency between the one-forward(single-step) trajectory and the multi-forward(multi-step) trajectory.
Accurate instantaneous velocity prediction serves as a stable anchor for distillation training and provides the exact real score required by the final one-step DMD stage(see \cref{sec:TAR}). In parallel, the self-consistent path-integral component aligns single-forward and multi-forward trajectories, supporting robust few-step sampling and facilitating subsequent one-step distillation.

Next, we obtain a one-step fake estimate by directly predicting the clean data from pure noise. Specifically, we sample $\epsilon \sim \mathcal{N}(0,I)$, and perform a single velocity prediction conditioned on the target timestep $t_{\text{tar}}$ as follows
\begin{gather}
    z^{\text{fake}} = \epsilon - v_{\theta}(\epsilon, t, z_{\text{lq}}, t_{\text{tar}}),
    \label{eq:generate_zfake}
\end{gather}
where $t=1,t_{\text{tar}}=0$ and $z_{\text{lq}}$ denotes the encoded LQ image.
To model the distribution of $z^{\text{fake}}$, we define a {fake trajectory} by corrupting the stopped-gradient $z^{\text{fake}}$ with Gaussian noise. Concretely, we sample $\epsilon^{\text{fake}} \sim \mathcal{N}(0,I)$ and $t' \sim \mathcal{U}(0,1)$, and form
\begin{gather}
    z^{\text{fake}}_{t'} = t'\,\epsilon^{\text{fake}} + (1-t')\,\mathrm{sg}(z^{\text{fake}}),
    \label{eq:zfake_forward} \\
    L_{\text{adv}}(\theta) = \mathbb{E}_{z_0,z_{\text{lq}}}
    \left\| v_{\theta}\!\left(z^{\text{fake}}_{t'},t',z_{\text{lq}},-t'\right) - \left(\epsilon^{\text{fake}} - \mathrm{sg}(z^{\text{fake}})\right) \right\|_2 ,
    \label{eq:loss_adv}
\end{gather}
where $\mathrm{sg}(\cdot)$ denotes stop-gradient. We optimize the fake trajectory solely with the flow-matching objective by making it conditioned on the corresponding current timestep and fake target timestep, implemented as $-t'$.

{Finally, the learned fake trajectory provides a {fake score} that can be contrasted with the {real score} from the real trajectory to perform DMD-based rectification. Concretely, we re-corrupt the one-step fake estimate $z^{\text{fake}}$ with gaussian noise and feed the perturbed latent into both trajectories to obtain paired real/fake scores. Their discrepancy yields a rectification signal that pulls the one-step prediction toward the real data distribution while pushing it away from the fake estimate distribution:
\begin{gather}
z^{\text{fake}}_{s} = s\,\epsilon^{\text{rectify}} + (1-s)\,\mathrm{sg}(z^{\text{fake}}),
\label{eq:loss_rectify_forward}\\
\Delta_v\!\left(z^{\text{fake}}_{s}\right) =
v_{\theta}\!\left(z^{\text{fake}}_{s}, s,z_{\text{lq}},-s\right) -
v_{\theta}\!\left(z^{\text{fake}}_{s}, s,z_{\text{lq}}, s\right),
\label{eq:loss_rectify_score}\\
\begin{aligned}
L_{\text{rectify}}&(\theta) = \\
\mathbb{E}_{z_0,z_{\text{lq}}}
\Big\| v_{\theta}\!\left(z_t,t,z_{\text{lq}},t_{\text{tar}}\right)
- \mathrm{sg}\!\Big[
v_{\theta}\!&\left(z_t,t,z_{\text{lq}},t_{\text{tar}}\right)
- \Delta_v\!\left(z^{\text{fake}}_{s}\right)
\Big] \Big\|_2
\end{aligned}
\label{eq:loss_rectify_dmd}
\end{gather}
where $t=1, t_{\text{tar}}=0$, $s\sim\mathcal{U}(0,1), \epsilon^{\text{rectify}}\sim\mathcal{N}(0,I)$ and $z_t = \epsilon$ given by \cref{eq:generate_zfake}. Overall, optimizing the fake-trajectory objective encourages the model to represent the distribution of the fake estimate $z^\text{fake}$, while the DMD rectification explicitly guides the one-step output $z^\text{fake}$ closer to the real data distribution and away from the fake one, establishing a self-adversarial training scheme.}

\textbf{Pixel-space fidelity anchoring.} {In super-resolution, a reconstruction objective is typically imposed to ensure fidelity to the ground-truth HQ image. Moreover, at the early stage of training, the model often fails to produce a meaningful one-step estimate $z^{\text{fake}}$, which in turn makes the learning of the fake trajectory ineffective. To address this issue, we introduce a pixel-domain fidelity constraint on $z^{\text{fake}}$ by decoding it into image space and directly supervising it with the HQ target. Specifically, we compute a mixed reconstruction loss consisting of MSE and LPIPS:
\begin{gather}
    x^{\text{fake}} = \mathcal{D}(z^{\text{fake}}),
    \label{eq:decode_fake}\\
    L_{\text{data}}(\theta)
    = L_{\text{MSE}}\!\left(x^{\text{fake}}, x_0\right)
    + \lambda_{\text{lpips}}\, L_{\text{lpips}}\!\left(x^{\text{fake}}, x_0\right).
    \label{eq:loss_data}
\end{gather}
where $\mathcal{D}(\cdot)$ denotes the VAE decoder.
This pixel-domain constraint anchors the one-step prediction to the HR supervision, preventing early collapse of $z^{\text{fake}}$ and providing a reliable training signal that stabilizes subsequent fake-trajectory modeling and DMD rectification.}




\subsection{Time-Aware Rectification}
\label{sec:TAR}

In the DMD stage of self-adversarial distillation, the rectification term $\Delta_v(z^{\text{fake}}_{s})$ represents the velocity discrepancy between the real and fake trajectories. This term serves as the primary distillation gradient signal to refine the one-step estimate ${z}^{\text{fake}}$. However, we observe that the efficacy of this rectification is inherently coupled with the noise level $s$. Naively sampling $s \sim \mathcal{U}(0,1)$ and assigning uniform weights fails to achieve optimal convergence, as it overlooks the {varying reliability of guidance signals} across different temporal stages. While prior works~\cite{dong2025tsd, zhang2026time} have also observed this issue, they typically resort to computationally expensive multi-forward averaging or simply fixed timesteps, which may compromise the quality of the distilled model. 

Specifically, the precision of the estimated velocity discrepancy undergoes significant fluctuations as the noise level $s$ varies:

    \textbf{Noise-dominant Regime ($s \to 1$):} When $s$ is large, the latent $z^{\text{fake}}_{s}$ is dominated by stochastic noise, resulting in a low signal-to-noise ratio. In this region, the structural information of the noisy latent is heavily corrupted, causing the instantaneous velocities predicted by both real and fake trajectories to exhibit high variance. Consequently, the rectification signal in this regime suffers from high velocity ambiguity and prediction error, which can introduce suboptimal updates and slow convergence.
    
    \textbf{Data-dominant Regime ($s \to 0$):} Conversely, as $s$ approaches the clean data manifold, the latent state retains more identifiable structural features. This allows the model to yield more accurate and consistent velocity predictions. The resulting discrepancy $\Delta_v$ provides a high-accuracy guidance signal that is crucial for recovering sharp edges and intricate textures, thus aligning the estimate ${z}^{\text{fake}}$ more closely with real data distribution.

To provide the optimization process more reliable guidance, we propose a Timestep-Aware Rectification scheduler. Instead of treating all timesteps $s$ equally, TAR applies a time-dependent scaling factor to suppress the unreliable gradients from high-noise states and amplify the precise signals from low-noise timesteps. The rectified velocity discrepancy is reformulated as:
\begin{equation}
    \Delta_v(z^{\text{fake}}_{s}) \leftarrow w(s) \cdot \Delta_v(z^{\text{fake}}_{s}), \quad \text{where } w(s) = \frac{(1-s)^2}{s}.
\end{equation}
By incorporating the weight $w(s)$, the training objective prioritizes timesteps with higher guidance reliability. This strategy effectively mitigates the misguidance caused by noise-dominant signals, facilitating the generation of high-frequency details, reducing one-step artifacts while ensuring the global structural integrity of the one-step estimate $z^\text{fake}$.


\subsection{Decoupled Timestep Condition}
\label{sec:DTC}

\textbf{Dual branch conditioning.} {DiT models operate on a tokenized representation of the noisy latent, which naturally enables spatial conditioning by concatenating the condition tokens with the noisy-latent tokens. The subsequent self-attention layers allow the noisy-latent tokens to attend to the condition tokens, thereby injecting conditional information in a spatially aligned manner. Following OminiControl~\cite{tan2025ominicontrol2efficientconditioningdiffusion}, we adopt a prior-preserved dual-branch conditioning design: we freeze the parameters associated with the first half of the token sequence (the {noise branch}) to preserve the pretrained generative prior, and only finetune the second half (the {condition branch}) to efficiently learn SR-specific conditioning. }

\textbf{Decoupled Timestep Condition.}
{Prior consistency-style methods (e.g., Shortcut, Meanflow) typically need two timesteps to represent $t_{\text{cur}}$ and $t_{\text{tar}}$. To let the model be aware of these two timesteps, they introduce the $t_{\text{tar}}$ by fusing it with the $t_{\text{cur}}$, either via numerical adjustment (e.g., $t \leftarrow t_{\text{cur}} - \tfrac{t_{\text{tar}}}{2}$) or by summing timestep embeddings (e.g., $\mathrm{temb} \leftarrow \mathrm{temb}_1(t_{\text{cur}}) + \mathrm{temb}_2(t_{\text{tar}})$). Such fusion will entangle the two different timestep signals, thus losing fine-grained noise-level information and target information, which is particularly harmful for super-resolution where fidelity is highly sensitive to the diffusion noise level.}

{We instead decouple the two timestep conditions: the noise branch is conditioned on the current timestep $t_{\text{cur}}$(noise level), while the condition branch is conditioned on the target timestep $t_{\text{tar}}$(destination of the flow step). This branch-wise injection preserves both timestep signals explicitly through attention, allowing the model to jointly reason about the current noise state and the intended denoising target without entangling them into a single embedding.}

\begin{algorithm}[tbp]
\caption{Any-Step Sampling of TEASR}
\label{alg:sampling}
\KwIn{Input LQ data $\boldsymbol{x}_{\text{lq}}$, Inference steps $N$, VAE encoder $E$ and Decoder $D$, model $\boldsymbol{v}_\theta$}
\KwOut{Restored result $\boldsymbol{x}_{\text{pred}}$}
\DontPrintSemicolon

$\boldsymbol{z}_{\text{lq}} \leftarrow {E}(\boldsymbol{x}_{\text{lq}})$\ 

Sample Gaussian noise $\boldsymbol{\epsilon} \sim \mathcal{N}(\mathbf{0}, \mathbf{I})$\;
$\boldsymbol{z}_{\tau_0} \leftarrow \boldsymbol{\epsilon}$\;

Construct time steps $\{\tau_0, \tau_1, \dots, \tau_N\}$ linearly from $1$ to $0$\;
$\Delta \tau \leftarrow 1 / N$\;

\For{$i \leftarrow 0$ \KwTo $N-1$}{
    
    $\boldsymbol{v}_{\text{pred}} \leftarrow \boldsymbol{v}_\theta(\underbrace{\boldsymbol{z}_{\tau_{i}}, \tau_{i}}_{\text{Noise Branch}}, \underbrace{\boldsymbol{z}_{\text{lq}}, \tau_{i+1}}_{\text{Cond Branch}})$\;
    
    $\boldsymbol{z}_{\tau_{i+1}} \leftarrow \boldsymbol{z}_{\tau_i} - \boldsymbol{v}_{\text{pred}} \cdot \Delta \tau$\;
}

$\boldsymbol{x}_{\text{pred}} \leftarrow {D}(\boldsymbol{z}_{\tau_N})$\;

\Return{$\boldsymbol{x}_{\text{pred}}$}
\end{algorithm}

\subsection{Any-Step Sampling}

The proposed TEASR framework, by virtue of its SAD, DTC and TAR, overcomes the limitations of previous fixed-step SR methods. It facilitates {any-step sampling}, allowing users to dynamically navigate the trade-off between inference efficiency and restoration quality at test time without retraining at different timesteps.

As detailed in \cref{alg:sampling}, the sampling procedure is formulated as a discrete denoising along the learned ODE trajectory. The core of our iterative refinement lies in the interaction between the dual-branch with the DTC design. For each step $i \in \{0, \dots, N-1\}$, the model $\boldsymbol{v}_\theta$ performs {co-reasoning} by simultaneously receiving two decoupled timestep signals of $t_{\text{cur}}$ and $t_{\text{tar}}$.

\section{Experiments}

\begin{table}[!t]
\centering
\caption{Quantitative comparison with state-of-the-art methods on both synthetic and real-world benchmarks. "Models" denotes the number of maintained models during training. The best and second best results of each metric are highlighted in \textcolor{red}{red} and \textcolor{blue}{blue}, respectively.}
\label{tab:main_comparison}
\resizebox{\textwidth}{!}{%
\begin{tabular}{l l c c ccccc cccc}
\toprule
{Datasets} & {Methods} & {Step(s)} & {Model(s)} & {PSNR$\uparrow$} & {SSIM$\uparrow$} & {LPIPS$\downarrow$} & {DISTS$\downarrow$} & {FID$\downarrow$} & {MUSIQ$\uparrow$} & {MANIQA$\uparrow$} & {NIQE$\downarrow$} \\
\midrule
& StableSR & 200 & 1 & 24.70 & 0.7085 & 0.3018& 0.2288& 128.51& 65.78& 0.6221& 5.91\\
& DiffBIR &50 & 1 & 24.75& 0.6567& 0.3636& 0.2312& 128.99& 64.98& 0.6246& 5.53\\
& SUPIR & 50& 1 & 25.03& 0.6705& 0.3749& 0.2510& 120.97& 57.86& 0.5656& 7.79\\
RealSR & DiT4SR & 40& 1 & 23.50& 0.6657& 0.3215& 0.2251& 118.55& 67.77& \textcolor{red}{0.6564}& 5.98\\
& OSEDiff & 1 & 3 & \textcolor{red}{25.15} & \textcolor{red}{0.7341} & 0.2921 & 0.2128 & 123.49 & \textcolor{blue}{69.09} & 0.6335 & 5.65\\
 & TSD-SR & 1 & 3& 24.81& 0.7172& \textcolor{blue}{0.2743}& \textcolor{blue}{0.2105}& \textcolor{blue}{114.48}& \textcolor{red}{71.18}& \textcolor{blue}{0.6345}& \textcolor{red}{5.12}\\
 & \textbf{TEASR} & 1 & 1 & \textcolor{blue}{24.83} & \textcolor{blue}{0.7198} & \textcolor{red}{0.2542} & \textcolor{red}{0.2003} & \textcolor{red}{103.97} & 68.50 & 0.6265 & \textcolor{blue}{5.43}\\
\midrule
 & StableSR & 200 & 1 & \textcolor{blue}{28.03} & 0.7536 & 0.3284& 0.2269& 148.98& 58.51& 0.5601& 6.52\\
 & DiffBIR &50 & 1 & 26.71& 0.6571& 0.4557& 0.2748& 166.79& 61.07& \textcolor{blue}{0.5930}& 6.31\\
 & SUPIR & 50& 1 & 26.69& 0.6655& 0.4328& 0.2767& 152.69& 55.47& 0.5339& 8.88\\
DrealSR & DiT4SR & 40& 1 & 25.40& 0.6657& 0.3869& 0.2508& 163.05& \textcolor{blue}{65.75}& \textcolor{red}{0.6287}& 6.92\\
& OSEDiff & 1 & 3 & \textcolor{blue}{27.92} & \textcolor{blue}{0.7835} & 0.2968 & 0.2165 & 135.29 & 64.65 & 0.5895 & 6.49\\
 & TSD-SR & 1 & 3& 27.77& \textcolor{black}{0.7559}& \textcolor{blue}{0.2966}& \textcolor{blue}{0.2136}& \textcolor{blue}{135.32}& \textcolor{red}{66.61}& 0.5850& \textcolor{red}{5.93}\\
 & \textbf{TEASR} & 1 & 1 & \textcolor{red}{28.54} & \textcolor{red}{0.7876} & \textcolor{red}{0.2722} & \textcolor{red}{0.2088} & \textcolor{red}{129.06} & 62.70 & 0.5582 & \textcolor{blue}{6.59}\\
\midrule
& StableSR & 200 & 1 & 23.26 & 0.5726 & 0.3113& 0.2048& 24.44& 65.92& 0.6192& 4.76\\
& DiffBIR &50 & 1 & {23.64}& 0.5647& 0.3524& 0.2128& 30.72& 65.81& 0.6210& \textcolor{blue}{4.70}\\
& SUPIR & 50& 1 & 23.15& 0.5436& 0.3632& 0.2256& 27.99& 62.61& 0.5938& 6.32\\
DIV2K-Val & DiT4SR & 40& 1 & 21.78& 0.5485& 0.3448& 0.2100& 30.57& \textcolor{blue}{68.07}& \textcolor{red}{0.6430}& 4.93\\
& OSEDiff & 1 & 3 & \textcolor{blue}{23.72} & \textcolor{blue}{0.6109} & 0.2942 & 0.1975 & 26.34 & 67.96 & 0.6147 & 4.71 \\
 & TSD-SR & 1 & 3& 23.02& 0.5808& \textcolor{red}{0.2673}& \textcolor{red}{0.1821}& \textcolor{red}{23.29}& \textcolor{red}{71.69}& \textcolor{blue}{0.6226}& \textcolor{red}{4.32}\\
 & \textbf{TEASR} & 1 & 1 & \textcolor{red}{24.36} & \textcolor{red}{0.6144} & \textcolor{blue}{0.2740} & \textcolor{blue}{0.1865} & \textcolor{blue}{23.40} & 65.72 & 0.5737 & 5.12\\
\bottomrule
\end{tabular}%
} 
\end{table}

\subsection{Experimental Settings}
\textbf{Implementations.} We train TEASR for the $\times 4$ Real-ISR task based on Qwen-Image-2512. The training process takes $\sim$47,000 steps on 2 NVIDIA A100 GPUs with a batch size of 1 and a gradient accumulation steps of 2. We use the AdamW~\cite{loshchilov2017decoupled} optimizer with learning rate 5e-5, $\beta$ =(0.9, 0.99), weight decay=0. We add LoRA~\cite{hu2022lora} to the condition branch with rank and alpha of 128. The noise branch is frozen. We set gradient clipping to 1.0.

\textbf{Datasets.} For training, we utilize LSDIR~\cite{li2023lsdir} and the first 10,000 images of FFHQ~\cite{karras2019style}, reaching a total of 94,991 images, where LQs are synthesized online following common degradation pipeline proposed in Real-ESRGAN~\cite{wang2021real} with the same parameter settings as previous works. For Real-ISR evaluation, we adopt two real-world datasets: RealSR~\cite{cai2019toward} and DRealSR~\cite{wei2020component}, and a synthetic dataset DIV2K-Val~\cite{agustsson2017ntire} following OSEDiff~\cite{wu2024one}. Without specific instruction, our model's prompt is left empty, other methods retain their default settings. 

\textbf{Metrics.} Both full-reference(FR) and no-reference(NR) metrics are employed for evaluation. FR metrics include pixel-level metrics PSNR and SSIM~\cite{ssim}, perceptual-level metrics LPIPS~\cite{lpips} and DISTS~\cite{dists} and distribution-level FID~\cite{fid}. NR image quality metrics include IQA methods MUSIQ~\cite{musiq}, MANIQA~\cite{maniqa} and NIQE~\cite{niqe}.

\textbf{Compared Methods.} We compare TEASR with SOTA diffusion-based one-step methods OSEDiff\cite{wu2024one}, TSD-SR~\cite{dong2025tsd}, multi-step methods StableSR~\cite{kim2021exploiting}, DiffBIR~\cite{lin2024diffbir}, SUPIR~\cite{yu2024scaling} and DiT4SR~\cite{duan2025dit4sr}. We obtain all results using their official open-source codes and models under their default settings unless otherwise specified.

\begin{figure}[!t]
    \centering
    \includegraphics[width=1.0\linewidth]{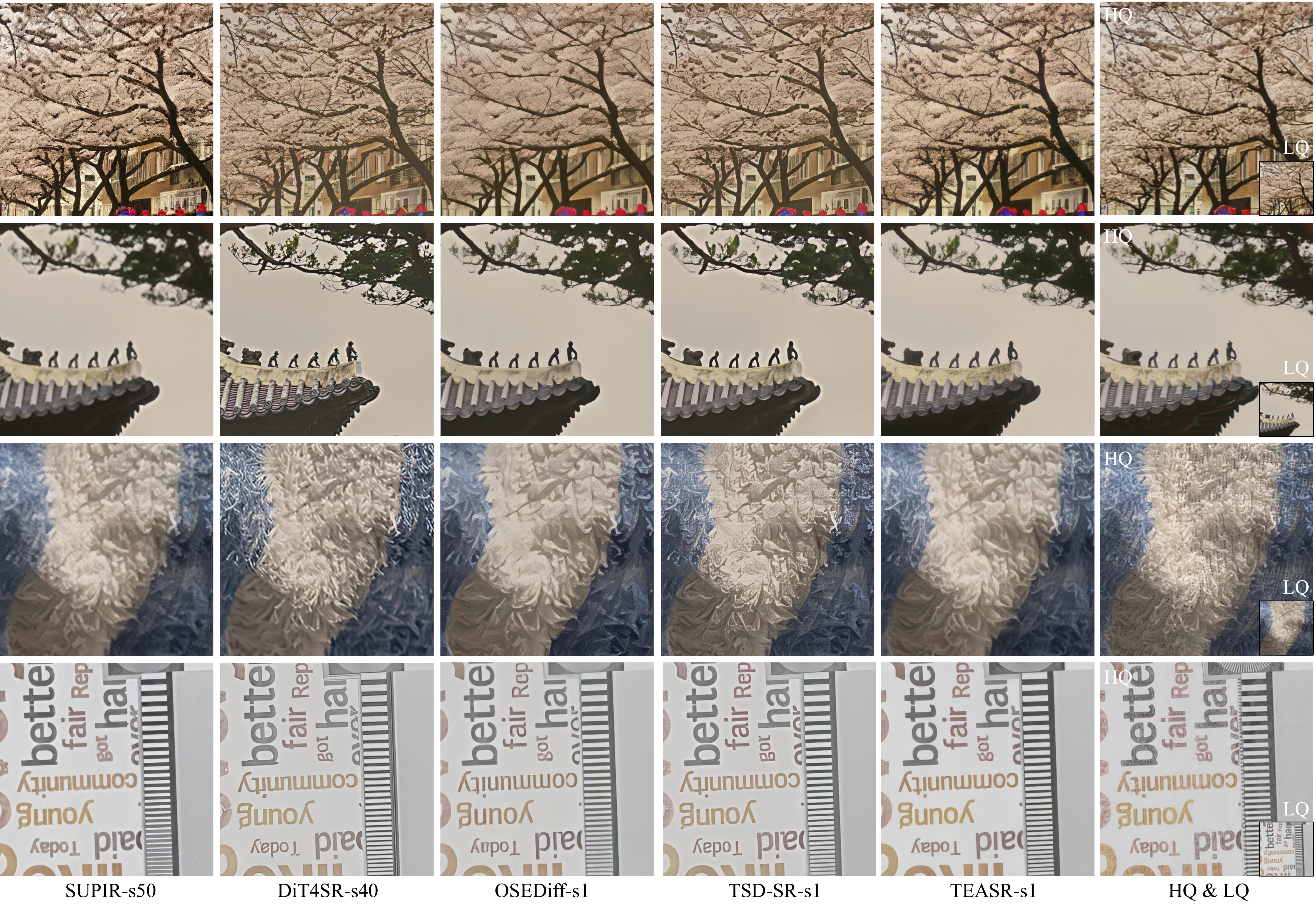}
    \caption{Qualitative comparison with state-of-the-art methods. 's' denotes the sampling steps. Compared with other state-of-the-art methods, TEASR produces the most natural outputs and the closest match to the ground truth.}
    \label{fig:main}
\end{figure}

\subsection{Comparison with State-of-the-Arts}
{\textbf{Quantitative Comparisons.}To validate the effectiveness of our method, we compare TEASR against state-of-the-art multi-step and one-step methods on both synthetic and real-world benchmarks. As shown in \cref{tab:main_comparison}, TEASR achieves an outstanding performance on restoration fidelity. Although OSEDiff obtains a slightly higher PSNR and SSIM on RealSR, it comes at the heavy memory consumption of maintaining three separate models during training. In contrast, our TEASR relies on only a single trainable model while securing the best perceptual metrics on RealSR, including the lowest LPIPS (\textbf{0.2542}) and FID (\textbf{103.97}). Furthermore, our method demonstrates superior generalization on DrealSR and DIV2K-Val, directly outperforming OSEDiff in PSNR (\textbf{28.54} vs. 27.92, and \textbf{24.36} vs. 23.72, respectively) and achieving the highest SSIM (\textbf{0.7876} and \textbf{0.6144}), while also demonstrating competitive performance in NR image quality metrics like NIQE. This comprehensive improvement indicates that our DTC precisely injects decoupled timestep information into the dual-branch DiT architecture, ensuring exceptional restoration fidelity without compromising the efficiency of our unified SAD framework. }

{\textbf{Qualitative Comparisons.} \cref{fig:main} presents the visual comparisons of our method against state-of-the-art models on the Real-ISR datasets. As observed in the first three rows, TEASR consistently provides high-quality and natural restoration results across complex natural scenes and fine-grained textures. While other methods often struggle with over-smoothing or unnatural details, our approach effectively recovers intricate visual elements with high fidelity. Furthermore, the bottom row highlights our method's precision in text and structural restoration. The result generated by TEASR exhibits well-defined stroke structures and is completely free from halo artifacts, clearly demonstrating that TEASR is highly capable of reconstructing complex characters, confirming TEASR's superior perceptual realism.
}

\begin{figure}[t]
    \centering
    \includegraphics[width=1.0\linewidth]{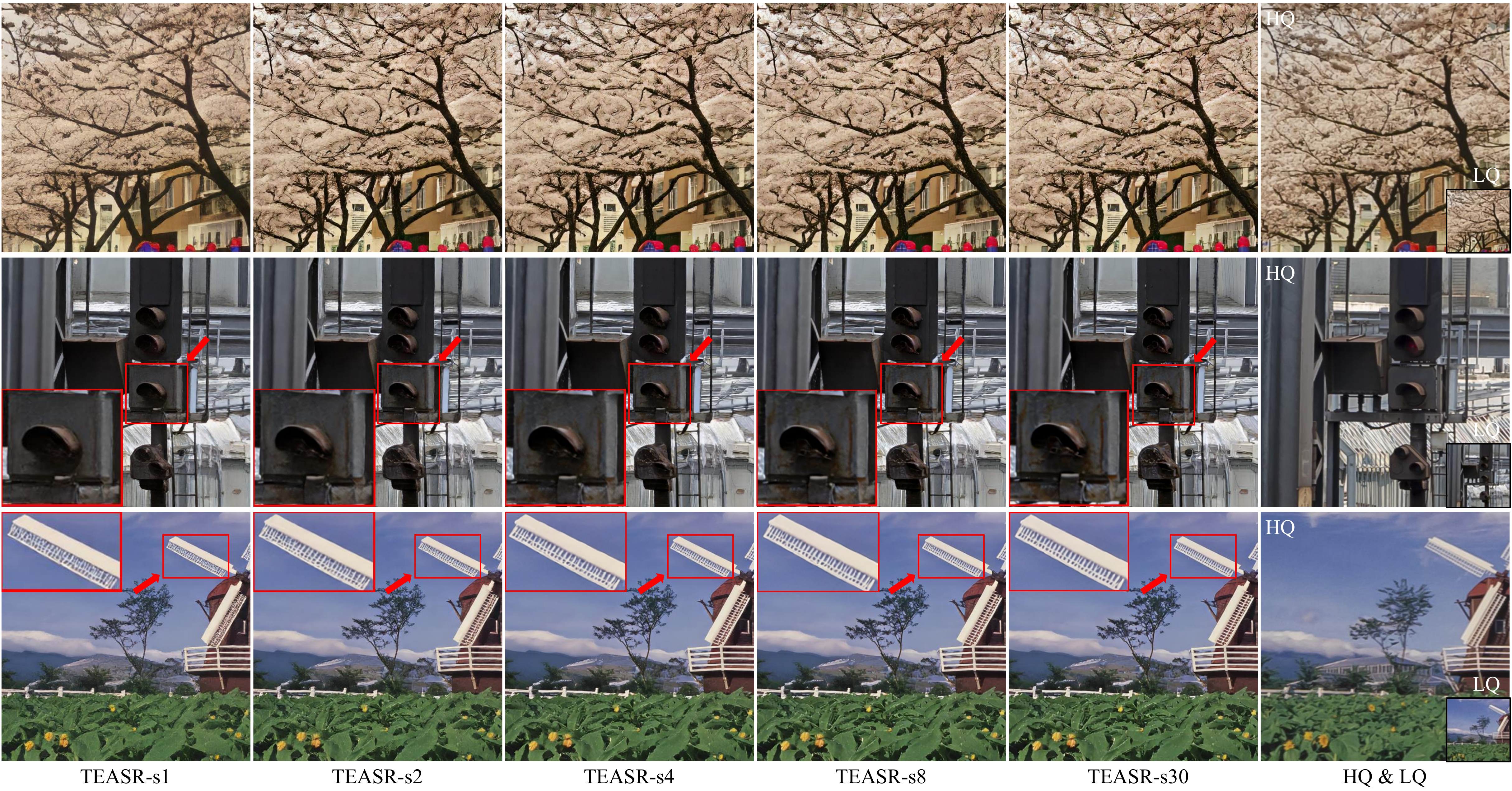}
    \caption{Qualitative comparison of TEASR across different sampling steps. 's' denotes the diffusion sampling steps. More sampling steps result in more natural and realistic outputs. Please zoom in for a better view.}
    \label{fig:anystep}
\end{figure}

\subsection{Validation of Training and Inference Efficiency}
\begin{figure}[t]
    \centering
    \includegraphics[width=0.95\linewidth]{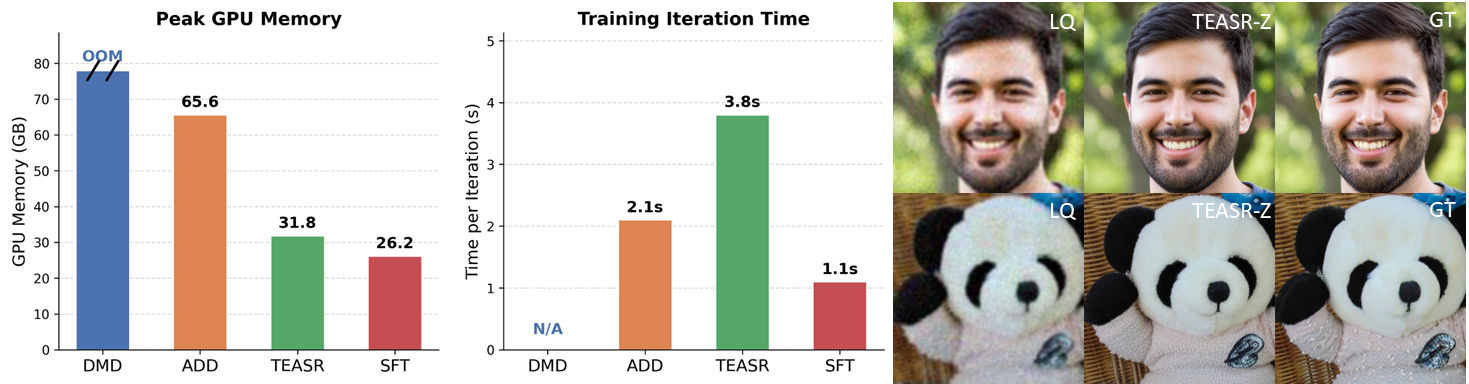}
    \caption{Comparisons of different distillation methods on Z-Image. The left panel illustrates the training memory consumption and time per iteration, while the right panel presents qualitative results generated by TEASR-Z.}
    \label{fig:z-image}
\end{figure}

To validate the training efficiency of our proposed method, we evaluate the training VRAM consumption and time per iteration of different distillation methods on the same backbone Z-Image-6B using a single NVIDIA A100 GPU. As shown in \cref{fig:z-image}, TEASR trained on Z-Image (denoted as TEASR-Z) achieves the lowest memory overhead while maintaining a reasonable time cost. Besides the right side is the visualizations of TEASR-Z, the results indicate that our framework can be applied to relatively small backbones, demonstrating our framework is robust to model scale.

\cref{tab:infer_efficiency} reports the inference efficiency comparisons on A100 GPU. Despite using the large backbone, TEASR still maintains practical inference speed while supporting one-step, few-step, multi-step SR within a unified model, achieving superior performance.

\begin{table}[t]
\scriptsize
\centering
\caption{Inference efficiency comparison.}
\resizebox{0.95\linewidth}{!}{
\begin{tabular}{l|c|c|c|c|c}
\toprule
Method & Backbone & Param. & Step(s) & Peak Mem. & Latency / Image\\
\midrule
OSEDiff & SD2.1 & 0.8B & 1 & 5934 MB & 0.15 s \\
DiT4SR & SD3 & 2B & 30 & {17244 MB} & {4.70 s}\\
TEASR & Qwen-Image & 20B & 1 / 4 / 30 & {59190 MB} & {0.72 / 2.48 / 10.31 s} \\
\bottomrule
\end{tabular}}
\label{tab:infer_efficiency}
\end{table}

\subsection{Any-Step Sampling}
Different from previous one-step SR methods, TEASR is able to perform any-step sampling, achieving better quality with more sampling steps. Specifically according to \cref{tab:any-step}, inference with more sampling steps results in a decrease in FR metrics while an increase in NR metrics. Visual comparison is shown in \cref{fig:anystep}. In the 2nd and 3rd row of \cref{fig:anystep}, increasing the number of steps will generate more details such as rust on the traffic light, make the texture of the windmill blades more straight and clear. To conclude, when increasing sampling steps, TEASR maintains good structural consistency, progressively generates more natural details and remove some unreal artifacts. More visual comparisons can be found in the \textbf{supplementary materials}.

\begin{table}[t]
\centering
\caption{Quantitative comparison across different sampling steps tested on RealSR of TEASR. The best and second best results of each metric are highlighted in \textcolor{red}{red} and \textcolor{blue}{blue}, respectively.}
\label{tab:any-step}
\resizebox{0.8\textwidth}{!}{%
\begin{tabular}{c ccccc cccc}
\toprule
{Step(s)} & {PSNR$\uparrow$} & {SSIM$\uparrow$} & {LPIPS$\downarrow$} & {DISTS$\downarrow$} & {FID$\downarrow$} & {MUSIQ$\uparrow$} & {MANIQA$\uparrow$} & {NIQE$\downarrow$}\\
\midrule
{1} & \textcolor{red}{24.83} & \textcolor{red}{0.7198} & \textcolor{red}{0.2542} & \textcolor{red}{0.2003} & \textcolor{red}{103.97} & 68.50 & 0.6265 & 5.4334\\
{2} & \textcolor{blue}{24.02} & \textcolor{blue}{0.6993} & 0.2707 & 0.2105 & 115.76 & \textcolor{blue}{68.65} & 0.6188 & 5.2770\\
{4} & 23.87& 0.6966& \textcolor{blue}{0.2704}& \textcolor{blue}{0.2085}& \textcolor{blue}{113.86}& \textcolor{red}{68.71}& \textcolor{red}{0.6345}& 5.3127\\
{8} & 23.69& 0.6892& 0.2849& 0.2169& 117.02& {68.52}& \textcolor{blue}{0.6321}& \textcolor{blue}{5.2344}\\
{30} & 23.61& 0.6837& 0.2913& 0.2183& 118.41& 68.42& 0.6305& \textcolor{red}{5.2327}\\
\bottomrule
\end{tabular}%
} 
\end{table}


\begin{figure}[t]
    \centering
    \includegraphics[width=0.8\linewidth]{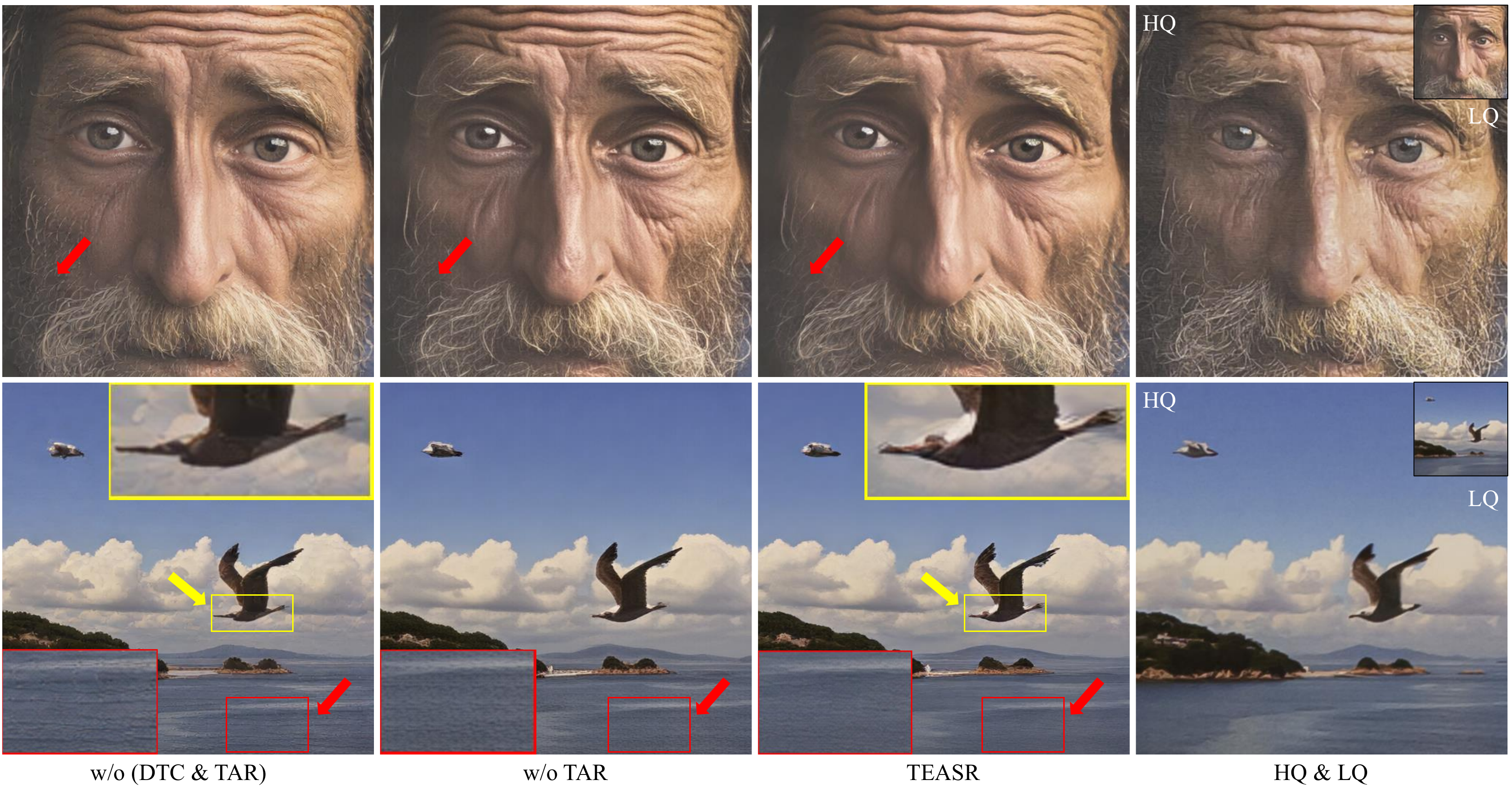}
    \caption{Qualitative comparison of ablation study. Please zoom in for a better view.}
    \label{fig:abla}
\end{figure}

\begin{table}[tbp]
\centering
\caption{Ablations. 'DTC' denotes Decoupled Timestep Condition. 'TAR' denotes Time-Aware Rectification. All results are sampled with one-step and tested on RealSR dataset.}
\label{tab:ablation}
\resizebox{0.9\textwidth}{!}{%
\begin{tabular}{cc ccccc ccc}
\toprule
\multicolumn{2}{c}{Structures} & \multirow{2}{*}{PSNR$\uparrow$} & \multirow{2}{*}{SSIM$\uparrow$} & \multirow{2}{*}{LPIPS$\downarrow$} & \multirow{2}{*}{DISTS$\downarrow$} & \multirow{2}{*}{FID$\downarrow$} & \multirow{2}{*}{MUSIQ$\uparrow$} & \multirow{2}{*}{MANIQA$\uparrow$} & \multirow{2}{*}{NIQE$\downarrow$}\\
\cmidrule(r){1-2}
DTC & TAR \\
\midrule
\ding{55} & \ding{55} & 24.93& 0.6840& 0.2904& 0.2317& 114.49& 68.26& 0.6288& 4.84\\
\ding{51} & \ding{55} & 25.52& 0.7329& 0.2335& 0.1994& 96.11& 66.39& 0.6252& 6.72\\
\ding{51} & \ding{51} & 24.83 & 0.7198 & {0.2542} & {0.2003} & {103.97} & 68.50 & 0.6265 & {5.43}\\
\bottomrule
\end{tabular}%
} 
\end{table}

\subsection{Ablation Study}

\textbf{Effectiveness of SAD.}
Self-Adversarial Distillation is our baseline for achieving few-step even one-step sampling. Without SAD, it will degenerate into a general multi-step model. So we do not conduct ablation study on it.

\textbf{Effectiveness of DTC.}
We explore the effectiveness of Decoupled Timestep Condition. Because DTC is a structural-level design which is not entangled with TAR(a loss-level design), we simply compare the baseline with the setting that only introduces dual-branch DiT with DTC. According to the 1st and 2nd row of \cref{tab:ablation}, DTC is superior to traditional timestep embedding fusion strategy in FR metrics, indicating that DTC helps model precisely restore most of the structural content on the image. In the 1st and 2nd columns of \cref{fig:abla}, results with DTC look more similar to the HQs. We consider it is due to the complete retention of both timesteps condition information, which can more correctly provide the model the information of noise level and target destination. This phenomenon may be more obvious in Real-ISR than in t2i generation.

\textbf{Effectiveness of TAR.}
We validate the effectiveness of Time-Aware Rectification, or, to elaborate, why is it necessary to add a time-aware weight to the rectification of the one-step generation process rather than use a constant weight. 
According to 2nd and 3rd row of \cref{tab:ablation}, the introduce of TAR results in a decrease in FR metrics while an increase in NR metrics. Visual differences are shown in 2nd and 3rd columns of \cref{fig:abla}, removing the Time-Aware Rectification results in the checkerboard-like artifact which is often observed in DiT architecture SR methods~\cite{li2025one, wu2025omgsr}. This artifact is a DiT-specific artifact happened in distilled methods when the generation path is not sufficiently optimized. TEASR successfully removes such artifact due to our TAR design provides a better guidance for the generation process.

\section{Discussions}
\textbf{Limitations.} While TEASR significantly reduces training overhead by maintaining only a single model, optimizing multiple composite objectives within a limited model capacity remains a challenge, which may lead to slight trade-offs in restoration quality under extreme conditions. Furthermore, although TEASR provides flexibility across different sampling steps, achieving fine-grained, independent control over fidelity and realism within a fixed step remains an open question for future exploration.

\textbf{Conclusions.} In this work, we present TEASR, a training-efficient any-step framework for Real-ISR. By introducing SAD, we eliminate the need for maintaining extra models, enabling large-scale DiT training with minimal resource requirements, significantly improving training efficiency. Coupled with our dual-branch design with DTC and TAR, TEASR offers remarkable sampling flexibility and superior restoration performance. We hope our efficient paradigm serves as a practical baseline for future research in Real-ISR.

\section{Acknowledgments}
This work is supported by National Natural Science Foundation of China (622713\\08,62571322), STCSM(24ZR1432000, 22DZ2229005), 111 plan (BP0719010),  and State Key Laboratory of UHD Video and Audio Production and Presentation.


%
%
\bibliographystyle{splncs04}
\bibliography{main}

\title{TEASR Supplementary} 

\titlerunning{TEASR.Supp.}

\author{Xiang Gao\orcidlink{0009-0003-6291-9522}, 
Chenxin Zhu\orcidlink{0009-0002-0679-6022}, 
Yushun Fang\orcidlink{0009-0008-8105-1083},\\ 
Qiang Hu\orcidlink{0000-0003-4645-9776}, 
Xiaoyun Zhang\orcidlink{0000-0001-7680-4062} \\[3pt] 
}

\authorrunning{X.~Gao et al.}

\institute{Shanghai Jiao Tong University, China \\
\email{\{frxg0918, Zcx-sjtu, ethanfang, qiang.hu, xiaoyun.zhang\}@sjtu.edu.cn}}

\maketitle

\section{Overview}

In this supplementary material, we provide additional analyses and implementation details for TEASR.
Specifically, we first report the training memory consumption of TEASR and compare it with representative diffusion distillation methods for super-resolution in Sec.~\ref{memory}.
We then present implementation details in Sec.~\ref{Implementation}, including the key hyperparameters, the complete training algorithm, and visualizations of the training losses.
Next, in Sec.~\ref{sec:derivation_rectify}, we provide a theoretical derivation of the proposed Time-Aware Rectification (TAR) loss.
Finally, in Sec.~\ref{visual}, we present additional qualitative results to further demonstrate the effectiveness of TEASR under different sampling steps.

\section{Training Memory Consumption}
\label{memory}


\cref{tab:memory_comparison} reports the training memory consumption of different distillation-based SR methods. 
Here, ``Baseline Memory'' denotes the memory usage of standard LoRA finetuning for t2i generation with the corresponding base model~\cite{esser2024scaling, rombach2022high, wu2025qwen}, while ``Proposed Method Memory'' denotes the memory usage of training the corresponding SR method. 
For all methods except TSD-SR~\cite{dong2025tsd}, memory is measured under the same setting: batch size 1, AdamW optimizer, image resolution 512*512, and gradient checkpointing enabled. 
For the baseline setting, the LoRA rank is matched to that used by the corresponding proposed method. 
The reported memory usage also includes the VAE and text encoder.

TSD-SR is a special case because it adopts data preprocessing to reduce training-time memory usage. 
According to community reproductions, when training TSD-SR with preprocessing, batch size 1, AdamW, and resolution 512*512, the default LoRA rank must be reduced from 64 to 16 in order to fit on a single 24GB RTX 3090 GPU.

\begin{table}[h]
\centering
\caption{Comparison of training memory consumption across different distillation-based SR methods. 
All results are measured on a single NVIDIA A100 GPU. 
$^*$ denotes results reported by community reproductions.}
\label{tab:memory_comparison}
\resizebox{0.9\textwidth}{!}{%
\begin{tabular}{lccccc}
\toprule
\textbf{Methods} & \textbf{Base} & \textbf{Params} & \textbf{\begin{tabular}[c]{@{}c@{}}Baseline \\Memory\end{tabular}} & \textbf{\begin{tabular}[c]{@{}c@{}}Proposed Method\\ Memory\end{tabular}} & \textbf{\begin{tabular}[c]{@{}c@{}}Memory\\ Increase\end{tabular}} \\ \midrule
OSEDiff & SD2.1 & 0.86 B & 4.62 GB & 22.87 GB & 395\% $\uparrow$ \\
PiSA-SR~\cite{sun2025pixel} & SD2.1 & 0.86 B & 4.62 GB & 15.06 GB & 226\% $\uparrow$ \\
TSD-SR$^*$ & SD3-m & 2 B & $\sim$10 GB & >24 GB & >140\% $\uparrow$ \\
TEASR & Qwen-Image & 20B & 67.18 GB & 77.69 GB & 15.6\% $\uparrow$ \\
\bottomrule
\end{tabular}%
} 
\end{table}

\section{Implementation Details}
\label{Implementation}

\textbf{Hyperparameter Overview. }We present all the main hyperparameters in \cref{tab:hyper}.
\begin{table}[!t]
\centering
\caption{A summary of the hyperparameters used in TEASR.}
\label{tab:hyper}
\resizebox{\textwidth}{!}{%
\begin{tabular}{l|c|c|c}
\toprule
Hyperparameter & Learning Rate & Batch Size & Optimizer
\\
Value & constant 5e-5 & 4 (gradient accumulation=2) & AdamW (beta=(0.9, 0.99))
\\
\midrule
Hyperparameter & LoRA Rank & LoRA Alpha & Training Iterations
\\
Value & 128 & 128 & 47,496
\\
\midrule
Hyperparameter & Training Resources & Training Precision & Gradient Checkpointing
\\
Value & 2*NVIDIA A100 GPUs & bfloat16 & True
\\
\bottomrule
\end{tabular}%
} 
\end{table}

\begin{figure}[!t]
    \centering
    \includegraphics[width=1.0\linewidth]{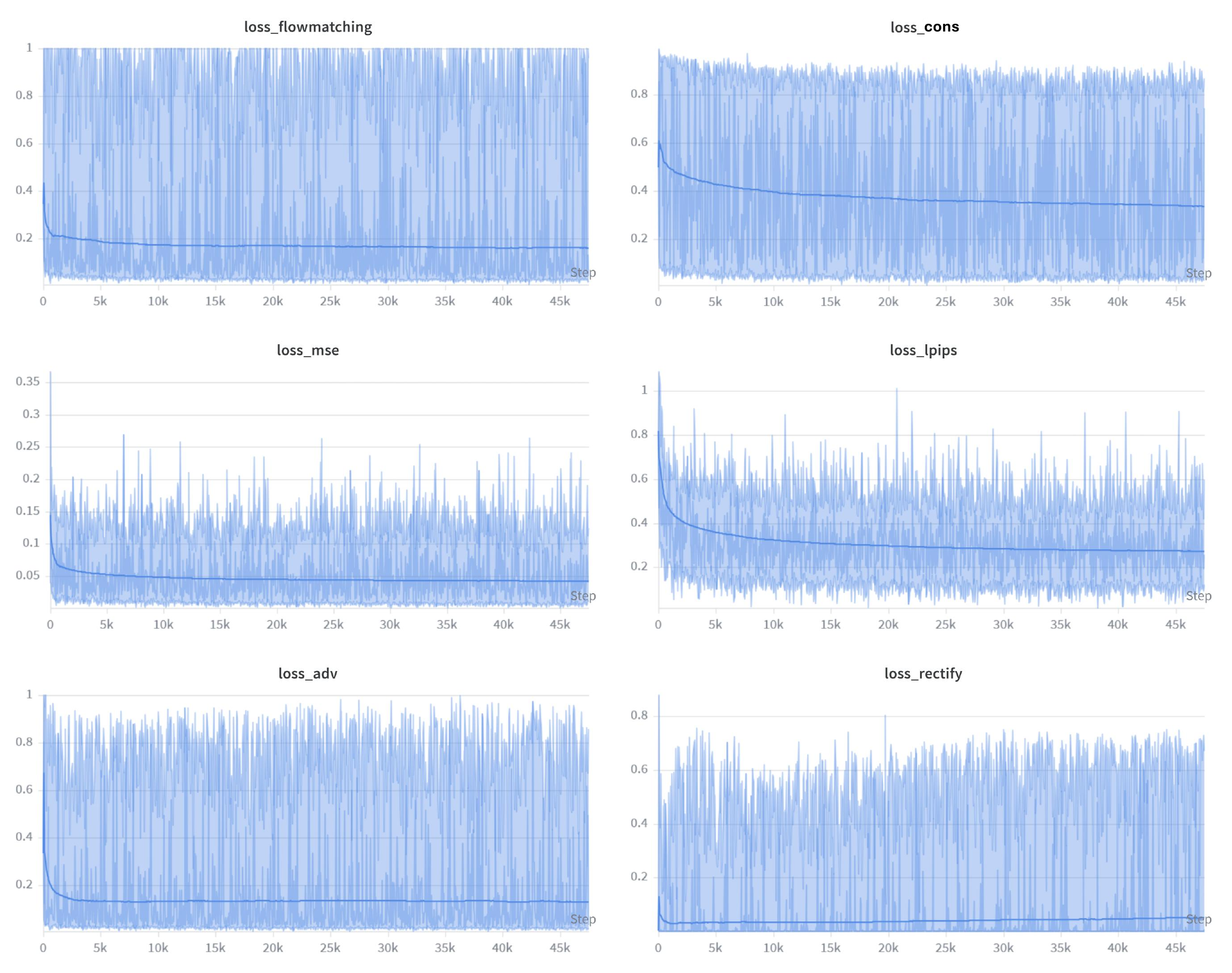}
    \caption{The visualization of the loss curves: the meaning of the titles corresponds to the symbols in the main paper. To avoid ambiguity, we manually correct the title of $L_{\text{cons}}$ to 'loss\_cons'.}
    \label{fig:loss_curve}
\end{figure}

\textbf{Loss Curve Visualization. }To help readers understand the training dynamics, we visualize the loss curves in \cref{fig:loss_curve}.  

\begin{algorithm}[!t]
\caption{Training Process of TEASR}
\label{alg:training}
\SetKwInOut{Input}{Input}\SetKwInOut{Output}{Output}
\Input{Training dataset $\mathcal{D} = \{({x}_{\text{hq}}, {x}_{\text{lq}})\}$, VAE encoder $E$, decoder $D$, trainable model $v_{\theta}$}
\DontPrintSemicolon

\While{not converged}{
    Sample a data batch $({x}_{\text{hq}}, {x}_{\text{lq}}) \sim \mathcal{D}$\;
    Extract latents: ${z}_{\text{hq}} \leftarrow E({x}_{\text{hq}})$, ${z}_{\text{lq}} \leftarrow E({x}_{\text{lq}})$\;
    
    \tcp{1. Real Trajectory Optimization (Flow Matching)}
    Sample current time $t_{\text{cur}} \sim \mathcal{U}(0, 1)$, target time $t_{\text{tar}} \sim \mathcal{U}(0, t_{\text{cur}})$, and noise ${\epsilon} \sim \mathcal{N}(\mathbf{0}, \mathbf{I})$\;
    ${z}_{\text{t}} \leftarrow t_{\text{cur}}{\epsilon}+(1-t_{\text{cur}}){z}_{\text{hq}}$\;
    ${L}_{\text{flow}} \leftarrow \| {v}_\theta({z}_{\text{t}}, t_{\text{cur}}, {z}_{\text{lq}}, t_{\text{tar}}) - ({\epsilon} - {z}_{\text{hq}}) \|_2^2$\;

    \tcp{2. Real Trajectory Optimization (Consistency)}
    Sample current time $t_{\text{cur}} \sim \mathcal{U}(0, 1)$, target time $t_{\text{tar}} \sim \mathcal{U}(0.95t_{\text{cur}}, t_{\text{cur}})$, and noise ${\epsilon} \sim \mathcal{N}(\mathbf{0}, \mathbf{I})$\;
    ${z}_{\text{t}} \leftarrow t_{\text{cur}}{\epsilon}+(1-t_{\text{cur}}){z}_{\text{hq}}$\;
    ${L}_{\text{cons}} \leftarrow L_{\text{RCGM}}({v}_\theta, {z}_{\text{t}}, t_{\text{cur}}, {z}_{\text{lq}}, t_{\text{tar}})$
    
    \tcp{3. One-Step Estimate \& Pixel-Space Anchoring}
    Sample pure noise ${\epsilon}^{\text{e2e}} \sim \mathcal{N}(\mathbf{0}, \mathbf{I})$\;
    ${v}_{\text{e2e}} \leftarrow {v}_\theta({\epsilon}^{\text{e2e}}, 1, {z}_{\text{lq}}, 0)$\;
    ${z}^{\text{fake}} \leftarrow {\epsilon}^{\text{e2e}} - {v}_{\text{e2e}}$\;
    ${x}^{\text{fake}} \leftarrow D({z}^{\text{fake}})$\;
    ${L}_{\text{data}} \leftarrow \text{MSE}({x}^{\text{fake}}, {x}_{\text{hq}}) + \lambda_{\text{lpips}}\text{LPIPS}({x}^{\text{fake}}, {x}_{\text{hq}})$\;
    
    \tcp{4. Fake Trajectory Optimization}
    Sample time $t' \sim \mathcal{U}(0, 1)$, noise ${\epsilon}^{\text{fake}} \sim \mathcal{N}(\mathbf{0}, \mathbf{I})$\;
    ${z}_{t'}^{\text{fake}} \leftarrow t'{\epsilon}^{\text{fake}}+(1-t')\text{sg}({z}^{\text{fake}})$\;
    ${L}_{\text{adv}} \leftarrow \| {v}_\theta({z}_{t'}^{\text{fake}}, t', {z}_{\text{lq}}, -t') - ({\epsilon}^{\text{fake}} - \text{sg}({z}^{\text{fake}})) \|_2^2$\;
    
    \tcp{5. Time-Aware Rectification (TAR)}
    Sample time $s \sim \mathcal{U}(0.05, 0.95)$, noise ${\epsilon}^{\text{rectify}} \sim \mathcal{N}(\mathbf{0}, \mathbf{I})$\;
    ${z}_s^{\text{fake}} \leftarrow s{\epsilon}^{\text{rectify}} + (1-s)\text{sg}({z}^{\text{fake}})$\;
    ${v}_{\text{fake}} \leftarrow {v}_\theta({z}_s^{\text{fake}}, s, {z}_{\text{lq}}, -s)$\; 
    ${v}_{\text{real}} \leftarrow {v}_\theta({z}_s^{\text{fake}}, s, {z}_{\text{lq}}, s)$ \;
    $\Delta_{{v}} \leftarrow {v}_{\text{fake}} - {v}_{\text{real}}$\;
    $w(s) \leftarrow (1-s)^2 / s$ \;
    ${v}_{\text{target}} \leftarrow \text{sg}({v}_{\text{e2e}} - w(s) \Delta_{{v}})$\;
    ${L}_{\text{rectify}} \leftarrow \| {v}_{\text{e2e}} - {v}_{\text{target}} \|_2^2$\;
    
    \tcp{6. Optimization Step}
    ${L}_{\text{total}} \leftarrow \lambda_{\text{flow}}{L}_{\text{flow}} + \lambda_{\text{cons}}{L}_{\text{cons}} + \lambda_{\text{data}}{L}_{\text{data}} + \lambda_{\text{adv}}{L}_{\text{adv}} + \lambda_{\text{rectify}}{L}_{\text{rectify}}$\;
    update $v_{\theta}$\;
}
\end{algorithm}

\textbf{Training Algorithm. }We provide \cref{alg:training} for better understanding of our SAD training scheme. 
Totally our loss is as
\begin{gather}
    L_{\text{total}} = \lambda_{\text{flow}}L_{\text{flow}} + \lambda_{\text{cons}}L_{\text{cons}} + \lambda_{\text{data}}L_{\text{data}} + 
    \lambda_{\text{adv}}L_{\text{adv}} + \lambda_{\text{rectify}}L_{\text{rectify}}.
\end{gather}

We simply set $\lambda_{\text{data}}=\lambda_{\text{adv}}=\lambda_{\text{rectify}}=1$. For $\lambda_{\text{flow}}$ and $\lambda_{\text{cons}}$, we make them aware of the distance between $t_{\text{cur}}$ and $t_{\text{tar}}$. Specifically, we calculate them as follows
\begin{gather}
    \lambda_{\text{flow}/\text{cons}}(t_{\text{cur}}, t_{\text{tar}}) = \tan((1-(t_{\text{cur}} - t_{\text{tar}}))\frac{\pi}{2.5}) + 1.
\end{gather}
This prioritizes the accuracy of local instantaneous velocity prediction, which is the base of multi-step diffusion model. We find this helpful to make the model converge rapidly to generate a reasonable result.  

\section{Theoretical Derivation of the Rectification Loss}
\label{sec:derivation_rectify}
In the main paper, we intuitively point out the motivation of TAR, which is also observed by TSD-SR. But they propose a DACM module to sample 4 different timesteps to get 4 distillation signals and assign them fixed weights, which significantly lower the training process. However our TAR design avoids the repetitive network forward in DACM. Here we analyze the design of TAR theoretically. 

The fundamental objective of our rectification mechanism is to align the distribution generated by our one-step model, denoted as $p_{\text{fake}}(z)$, with the real data distribution $p_{\text{real}}(z)$. Inspired by distribution matching principles, this alignment can be formulated as minimizing the Kullback-Leibler (KL) divergence over the intermediate noise levels $s \in (0, 1)$ as
\begin{equation}
    \min_{\theta} \mathbb{E}_{s \sim \mathcal{U}(0,1)} \left[ \mathcal{D}_{\text{KL}}\left( p_{\text{fake}}(z^{\text{fake}}_s) \parallel p_{\text{real}}(z^{\text{fake}}_s) \right) \right].
\end{equation}
Following DMD~\cite{yin2024one}, the gradient of this KL divergence with respect to the generated fake samples $z^{\text{fake}}$ can be derived from the score functions of the respective trajectories as
\begin{gather}
    \nabla_{\theta} \mathcal{D}_{\text{KL}} = - \nabla_{\theta}\mathbb{E}_{s \sim \mathcal{U}(0,1)}\left( \log p_{\text{real}}(z^{\text{fake}}_s) - \log p_{\text{fake}}(z^{\text{fake}}_s) \right)\\
    = \mathbb{E}_{s \sim \mathcal{U}(0,1)}(\nabla_{\theta}\log p_{\text{fake}}(z^{\text{fake}}_s) - \nabla_{\theta}\log p_{\text{real}}(z^{\text{fake}}_s))\\
    =\mathbb{E}_{s \sim \mathcal{U}(0,1)}(\nabla_{z^{\text{fake}}_s}\log p_{\text{fake}}(z^{\text{fake}}_s) - \nabla_{z^{\text{fake}}_s}\log p_{\text{real}}(z^{\text{fake}}_s))\frac{\partial z^{\text{fake}}_s}{\partial \theta}
\end{gather}
Within the flow matching framework, the score function $\nabla_{z_s} \log p(z_s)$ is inherently linked to the velocity field $v(z_s, s)$ as 
\begin{gather}
    \nabla_{z_s} \log p(z_s) = -\frac{z_s+(1-s)v_{\theta}(z_s,s,z_{\text{lq}},s)}{s}
\end{gather}

Consequently, the discrepancy between the score functions can be effectively reparameterized by the difference in their corresponding velocity fields. A key innovation in TEASR is the Decoupled Timestep Condition (DTC), which enables a single network $v_\theta$ to estimate both the real and fake velocities by simply altering the explicit target timestep condition. Specifically, the real trajectory is conditioned on $s$, while the one-step induced fake trajectory is identified by the negative timestep $-s$. Thus, the velocity error can be instantiated within our single-model framework as
\begin{equation}
    \Delta_v(z_s^{\text{fake}}) = v_\theta(z_s^{\text{fake}}, s, z_{\text{lq}}, -s) - v_\theta(z_s^{\text{fake}}, s, z_{\text{lq}}, s).
\end{equation}
Then the KL divergence can be derived as
\begin{gather}
    \nabla_{\theta} \mathcal{D}_{\text{KL}} = \\
    \mathbb{E}_{s \sim \mathcal{U}(0,1)}[-\frac{z^{\text{fake}}_s+(1-s)v^{\text{fake}}_{\theta}(z^{\text{fake}}_s)}{s}+\frac{z^{\text{fake}}_s+(1-s)v^{\text{real}}_{\theta}(z^{\text{fake}}_s)}{s}]\frac{\partial z^{\text{fake}}_s}{\partial \theta}\\
    =\mathbb{E}_{s \sim \mathcal{U}(0,1)}[-\frac{1-s}{s}(v_{\theta}(z^{\text{fake}}_s,s,z_{\text{lq}},-s)-v_{\theta}(z^{\text{fake}}_s,s,z_{\text{lq}},s))]\frac{\partial z^{\text{fake}}_s}{\partial \theta}\\
    =\mathbb{E}_{s \sim \mathcal{U}(0,1)}[-\frac{1-s}{s}\Delta_v(z^{\text{fake}}_s)]\frac{\partial z^{\text{fake}}_s}{\partial \theta}
\end{gather}
while the rightmost item is
\begin{gather}
    \frac{\partial z^{\text{fake}}_s}{\partial \theta}=\frac{\partial[s\epsilon^{\text{rectify}} + (1-s)z^{\text{fake}}]}{\partial \theta}=(1-s)\frac{\partial z^{\text{fake}}}{\partial \theta}\\
    =(1-s)\frac{\partial (\epsilon^{\text{e2e}}-v_\theta(\epsilon^{\text{e2e}}, 1, z_{\text{lq}},0))}{\partial \theta}\\
    =-(1-s)\frac{\partial v_\theta(\epsilon^{\text{e2e}}, 1, z_{\text{lq}},0)}{\partial \theta}
\end{gather}
Then the final KL divergence is 
\begin{gather}
    \nabla_{\theta} \mathcal{D}_{\text{KL}} = \mathbb{E}_{s \sim \mathcal{U}(0,1)}[-\frac{1-s}{s}\Delta_v(z^{\text{fake}}_s)]\frac{\partial z^{\text{fake}}_s}{\partial \theta}\\
    =\mathbb{E}_{s \sim \mathcal{U}(0,1)}[\frac{(1-s)^2}{s}\Delta_v(z^{\text{fake}}_s)]\frac{\partial v_\theta(\epsilon^{\text{e2e}}, 1, z_{\text{lq}},0)}{\partial \theta}
\end{gather}
To better rectify the generation process, we adjust the initial one-step velocity prediction, denoted as $v_\theta(\epsilon^{\text{e2e}}, 1, z_{\text{lq}},0)$ by incorporating our Time-Aware Rectification weight $w(s)=\frac{(1-s)^2}{s}$, we obtain the optimal target velocity for the one-step prediction as
\begin{gather}
    v_{\text{target}} = \text{sg}\left[ v_\theta(\epsilon^{\text{e2e}}, 1, z_{\text{lq}},0) - w(s) \Delta_v(z_s^{\text{fake}}) \right],
\end{gather}

Finally, the rectification loss ${L}_{\text{rectify}}$ is computed as the mean squared error between the model's one-step velocity prediction and the constructed target velocity as
\begin{equation}
        L_{\text{rectify}}=||v_\theta(\epsilon^{\text{e2e}}, 1, z_{\text{lq}},0) - v_{\text{target}}||^2_2.
\end{equation}

\section{Clarification of any-step sampling}
\begin{figure}[h]
    \centering
    \includegraphics[width=0.95\linewidth]{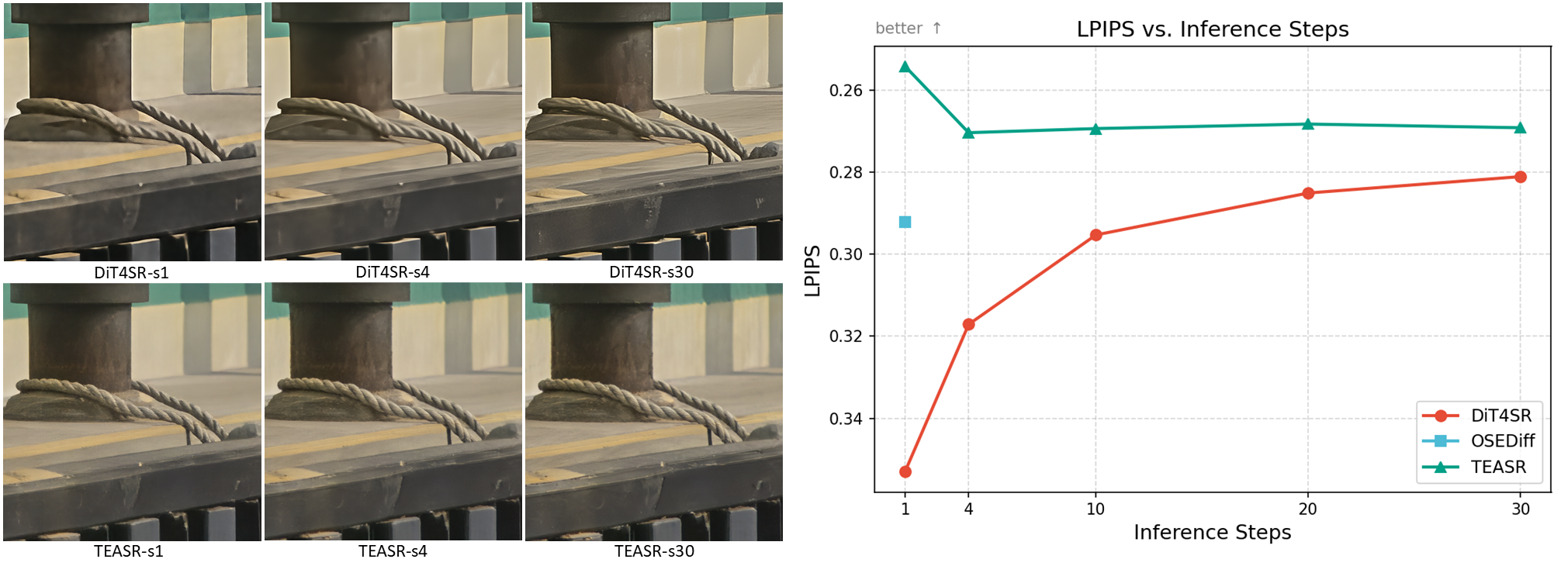}
    \caption{Comparisons of any-step sampling.}
    \label{fig:enter-label}
\end{figure}

Standard diffusion based SR models allow flexible sampling steps, but their one-step or few-step results are usually unsatisfactory. In contrast, DMD/ADD-based methods are optimized for fixed inference manner, limiting their ability to perform multi-step refinement. TEASR unifies one-step, few-step, and multi-step restoration within a single model while maintaining stable restoration quality across different sampling budgets. As shown in \cref{fig:enter-label}, DiT4SR \textbf{performs poorly} under one-step and few-step sampling, whereas TEASR achieves \textbf{consistently strong results}. Additional visualizations are provided in \cref{visual}.

\section{TAR motivation \& sensitivity}
We provide full theoretical derivation in \cref{sec:derivation_rectify} for TAR. Under the flow-matching score relation, aligning $z^{\text{fake}}$ with $p_{\text{real}}$ naturally leads to the optimal weight $w(s)=\frac{(1-s)^2}{s}$.
We further include a sensitivity analysis with alternative schedules on RealSR in \cref{tab:tar_sensitivity}. Different schedules remain \textbf{stable} during training, while our final TAR weighting achieves the \textbf{best} perceptual quality and the \textbf{fewest} artifacts according to both NR metrics and human observation.

\begin{table}[h]

\centering
\caption{TAR sensitivity analysis on RealSR dataset.}
\resizebox{\linewidth}{!}{
\begin{tabular}{l|c|c|c|c|c}
\toprule
Schedule $w(s)$ & LPIPS$\downarrow$ & MUSIQ$\uparrow$ & MANIQA$\uparrow$ & NIQE$\downarrow$ & Artifact Observation \\
\midrule
$1$ & \textcolor{red}{0.2335} & {66.39} & {0.6252} & {6.72} & checkerboard artifacts \\
$1-s$ &{0.2408} & {67.37} & {0.6203} & {5.60} & {fewer artifacts} \\
$(1-s)^2/s$ & 0.2542 & \textcolor{red}{68.50} & \textcolor{red}{0.6265} & \textcolor{red}{5.43} & least artifacts \\
\bottomrule
\end{tabular}}
\label{tab:tar_sensitivity}
\end{table}

\section{Early training stability}
\begin{figure}[h]
    \centering
    \includegraphics[width=0.9\linewidth]{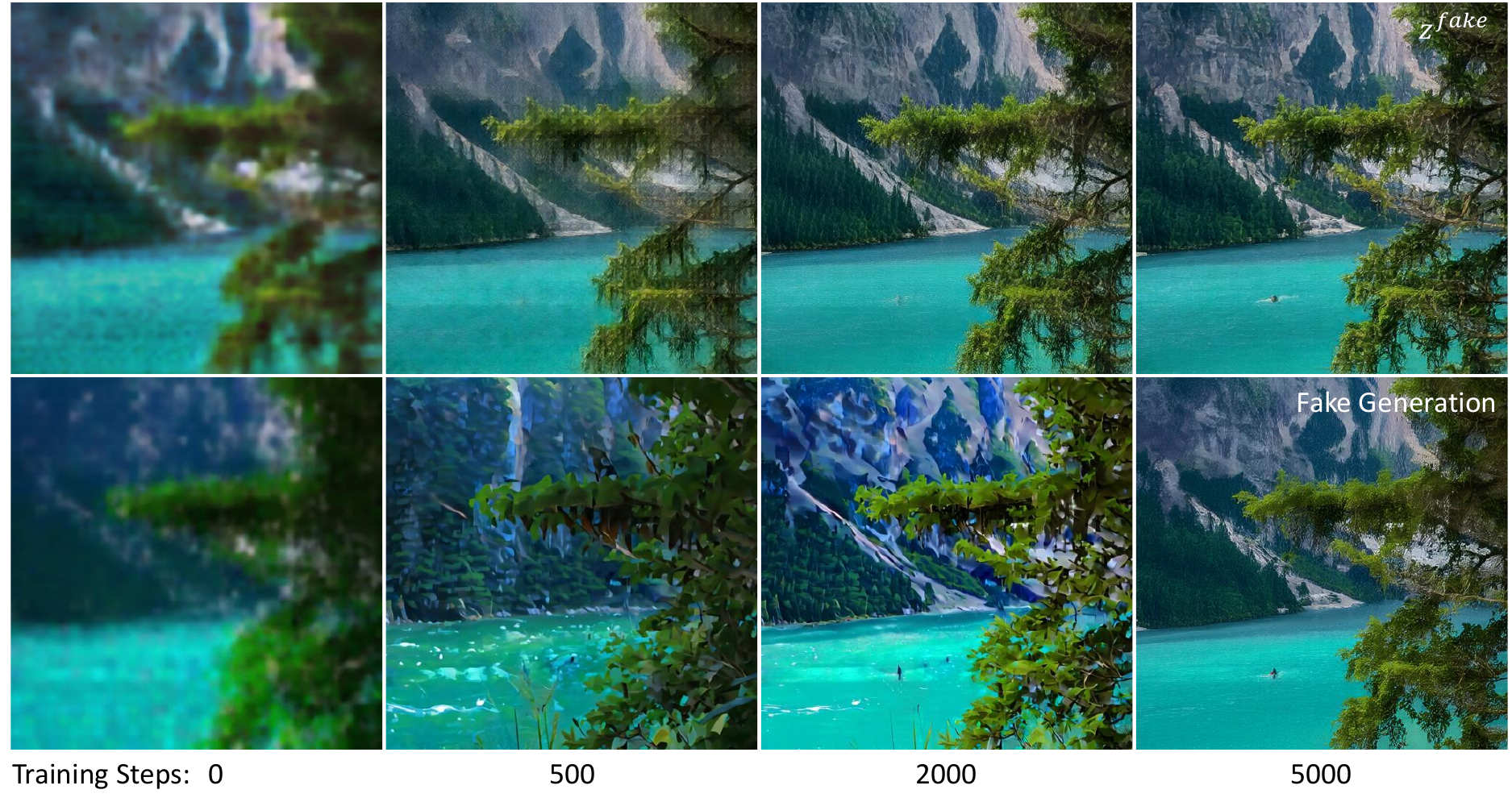}
    \phantomcaption
    \label{fig:SAD_visualization}
\end{figure}

We visualize both the one-step estimate $z^{\text{fake}}$ and the fake trajectory generation results with 30-steps sampling in \cref{fig:SAD_visualization}. During early training, both $z^{\text{fake}}$ and the fake trajectory rapidly captures the global structure but lacks realistic details. As training progresses, both become realistic stably, enabling the fake trajectory to provide increasingly informative adversarial guidance. The full training curves are provided in \cref{fig:loss_curve}, where all losses decrease smoothly and stably. Together, the visualizations and metrics demonstrate the \textbf{stability} of TEASR training.

\section{More Visual Results}
\label{visual}
\begin{figure}[H]
    \centering
    \includegraphics[width=0.9\linewidth]{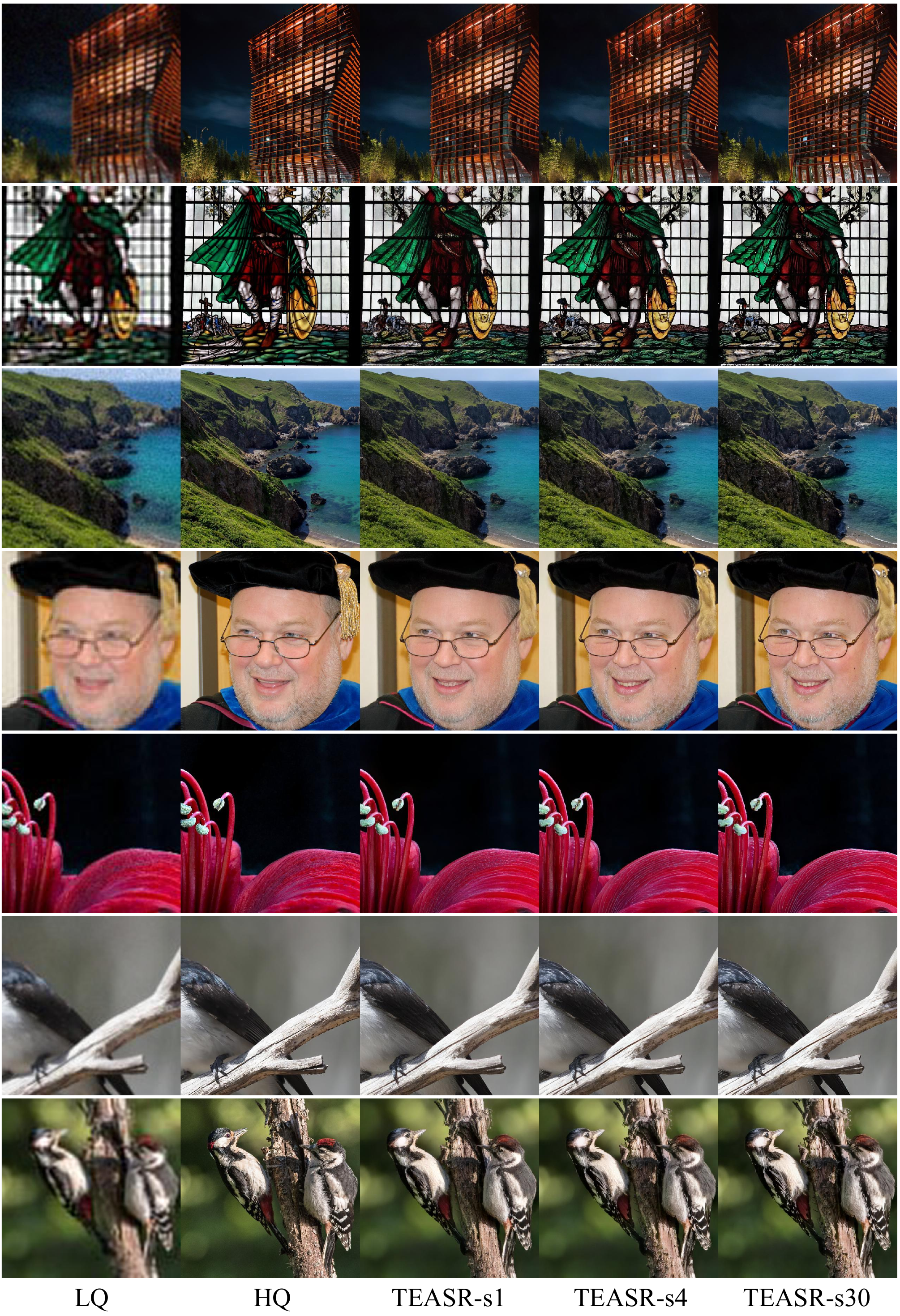}
    \caption{Additional qualitative comparisons of TEASR under different sampling steps. 
From left to right: low-quality input (LQ), ground truth (HQ), and TEASR results with 1, 4, and 30 sampling steps, respectively.}
    \label{fig:any_1}
\end{figure}

\begin{figure}[H]
    \centering
    \includegraphics[width=0.95\linewidth]{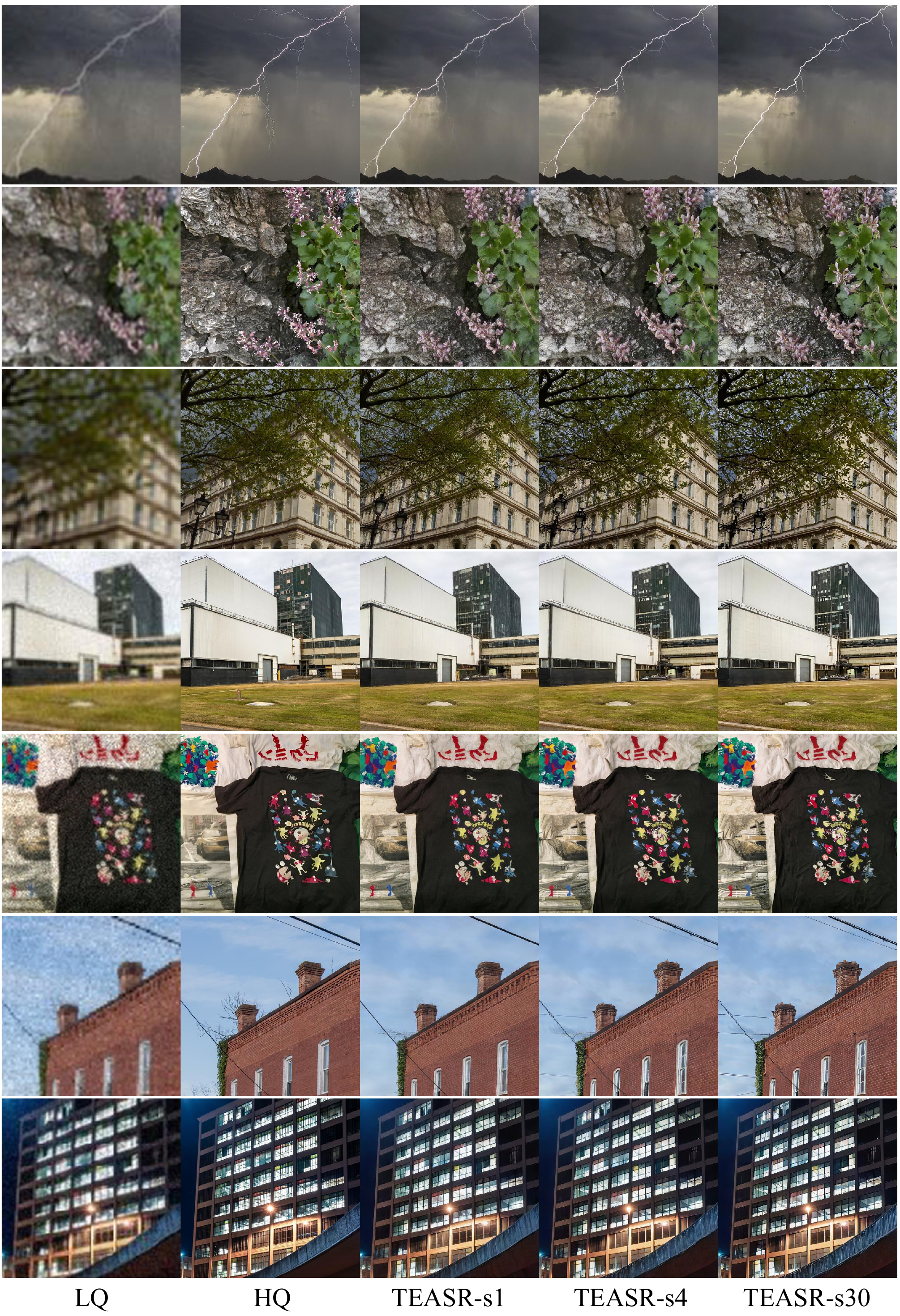}
    \caption{Additional qualitative comparisons of TEASR under different sampling steps. 
From left to right: low-quality input (LQ), ground truth (HQ), and TEASR results with 1, 4, and 30 sampling steps, respectively.}
    \label{fig:any_2}
\end{figure}

\begin{figure}[H]
    \centering
    \includegraphics[width=0.95\linewidth]{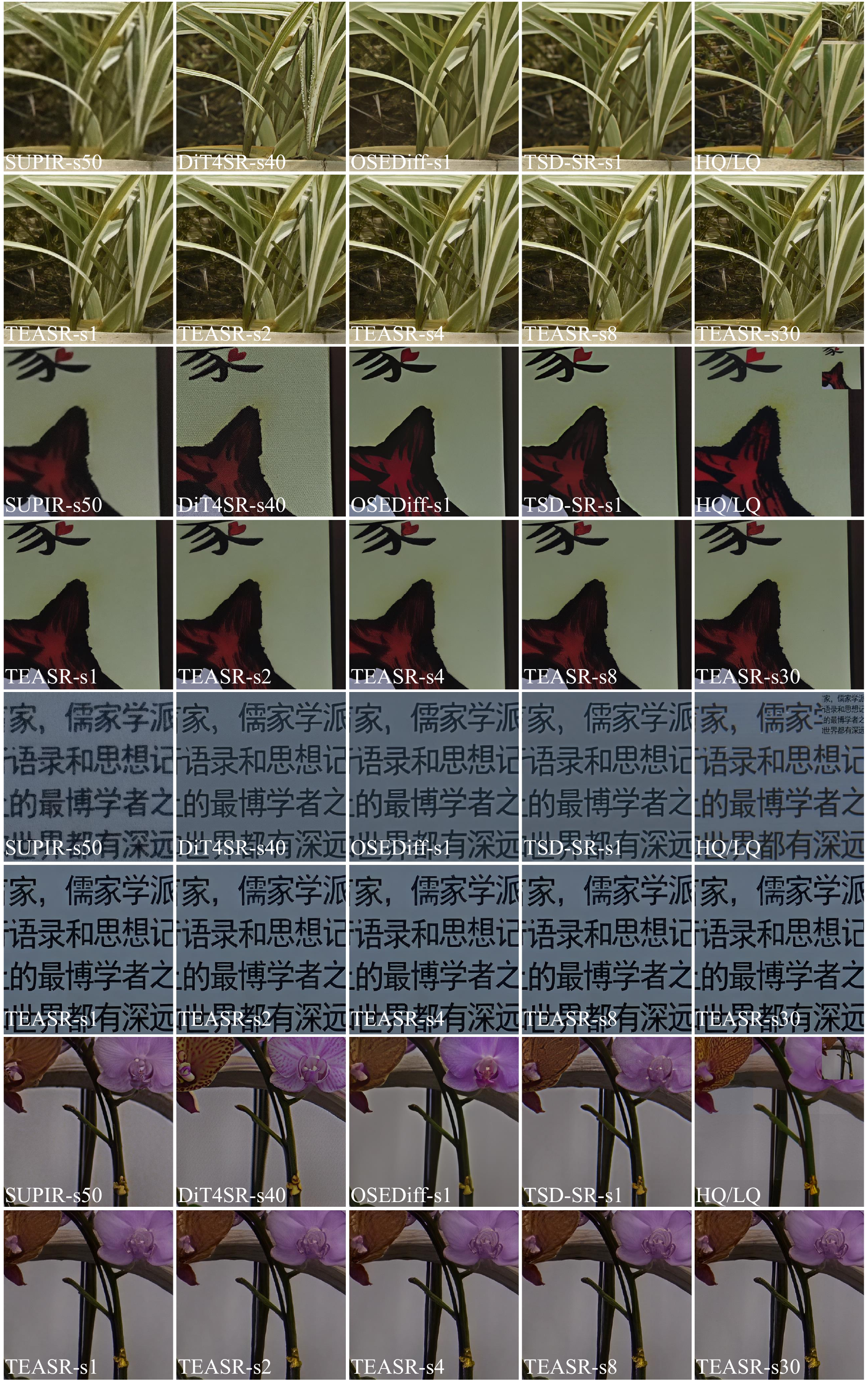}
    \caption{Additional qualitative comparisons of TEASR under different sampling steps and state-of-the-art SR methods.}
    \label{fig:compare_1}
\end{figure}

%
%


\end{document}